\documentclass{article} 
\usepackage{iclr2025_conference,times}

\usepackage{amsmath,amsfonts,bm}

\def\eqref#1{equation~\ref{#1}}

\def\1{\bm{1}}

\DeclareMathAlphabet{\mathsfit}{\encodingdefault}{\sfdefault}{m}{sl}
\SetMathAlphabet{\mathsfit}{bold}{\encodingdefault}{\sfdefault}{bx}{n}

\usepackage{hyperref}
\usepackage{url}

\usepackage{enumitem}
\usepackage{multirow}
\usepackage{graphicx}
\usepackage{pifont}
\usepackage{bbding}
\usepackage{subfigure}
\usepackage{booktabs}
\usepackage{amsmath} 
\usepackage{xcolor}
\usepackage{CJKutf8}
\usepackage{wrapfig}

\title{Benchmarking Multimodal Retrieval Augmented Generation with Dynamic VQA Dataset and Self-adaptive Planning Agent}

\author{Yangning Li\textsuperscript{\rm 1,2}\thanks{Equal Contribution. $^{\ddag}$Corresponding Author.}, Yinghui Li\textsuperscript{\rm 1*}, Xinyu Wang\textsuperscript{\rm 3\ddag}, Yong Jiang\textsuperscript{\rm 3\ddag}, Zhen Zhang\textsuperscript{\rm 3}, Xinran Zheng\textsuperscript{\rm 4}\\ \textbf{ Hui Wang\textsuperscript{\rm 2}, Hai-Tao Zheng\textsuperscript{\rm 1,2\ddag}, Philip S. Yu\textsuperscript{\rm 5}, Fei Huang\textsuperscript{\rm 3}, Jingren Zhou\textsuperscript{\rm 3}}\\
\textsuperscript{\rm 1}Shenzhen International Graduate School, Tsinghua University
\textsuperscript{\rm 2}Peng Cheng Laboratory \\\textsuperscript{\rm 3}Tongyi Lab, Alibaba Group
\textsuperscript{\rm 4}University College London \textsuperscript{\rm 5}University of Illinois Chicago
}

\iclrfinalcopy 
\begin{document}

\maketitle

\begin{abstract}
Multimodal Retrieval Augmented Generation (mRAG) plays an important role in mitigating the “hallucination” issue inherent in multimodal large language models (MLLMs). Although promising, existing heuristic mRAGs typically predefined fixed retrieval processes, which causes two issues: (1) Non-adaptive Retrieval Queries. (2) Overloaded Retrieval Queries. However, these flaws cannot be adequately reflected by current knowledge-seeking visual question answering (VQA) datasets, since the most required knowledge can be readily obtained with a standard two-step retrieval. To bridge the dataset gap, we first construct Dyn-VQA dataset, consisting of three types of ``dynamic'' questions, which require complex knowledge retrieval strategies variable in query, tool, and time: (1) Questions with rapidly changing answers. (2) Questions requiring multi-modal knowledge. (3) Multi-hop questions. Experiments on Dyn-VQA reveal that existing heuristic mRAGs struggle to provide sufficient and precisely relevant knowledge for dynamic questions due to their rigid retrieval processes. Hence, we further propose the first self-adaptive planning agent for multimodal retrieval, \textbf{OmniSearch}. The underlying idea is to emulate the human behavior in question solution which dynamically decomposes complex multimodal questions into sub-question chains with retrieval action. Extensive experiments\footnote{This work was led by Xinyu Wang at Alibaba Group, with Yangning, Yinghui, and Zheng Zhang as his interns. The code and dataset is open-sourced at \url{https://github.com/Alibaba-NLP/OmniSearch}.} prove the effectiveness of our OmniSearch, also provide direction for advancing mRAG.
\end{abstract}

\section{Introduction}
\begin{wrapfigure}{r}{0.5\textwidth}
\centering
\includegraphics[width=0.5\textwidth]{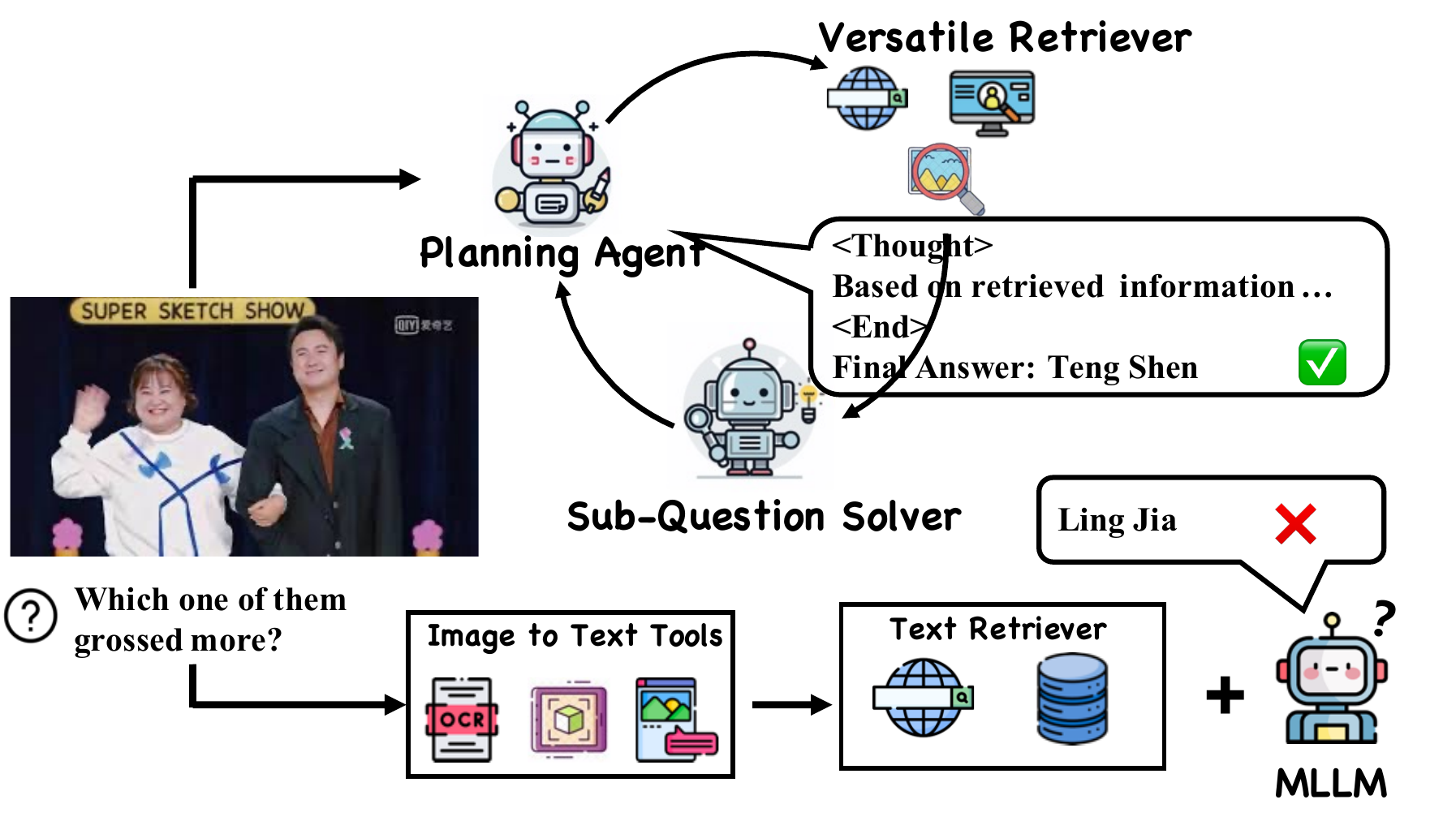}
\caption{Bottom: Heuristic mRAG based VQA. Upper: OmniSearch based VQA.}
\label{fig:intro_method}
\vspace{-1em}
\end{wrapfigure}
Multimodal Retrieval Augmented Generation (mRAG) \citep{zhao2024retrieval, zhao2023retrieving, gao2023retrieval}  aims to provide more comprehensive, accurate and up-to-date knowledge from external sources for AI systems. It has emerged as a key technology to mitigate the ``hallucination'' issue \citep{liu2024survey, bai2024hallucination} inherent in multimodal large language models (MLLMs). 

\begin{figure*}[]
\centering
\scalebox{0.75}{
\includegraphics[width=0.99\textwidth]{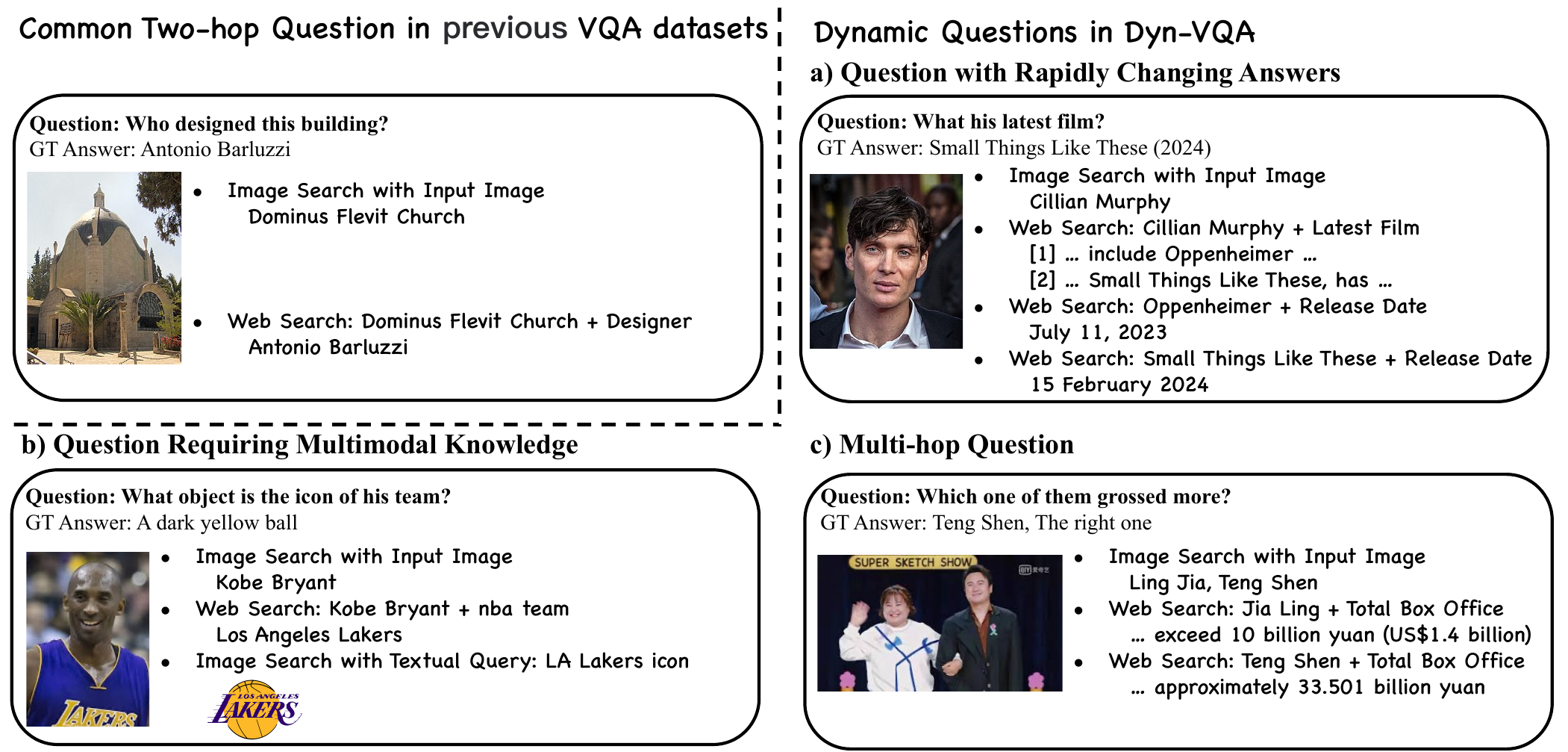}}
\vspace{-0.4cm}
\caption{Dynamic VQA examples that require different kinds of retrieval strategies.}
\label{fig:intro_data}
\vspace{-1.9em}
\end{figure*}

Existing heuristic mRAG methods typically predefined fixed retrieval processes that ground all modalities into one primary modality (usually text), then retrieve for a single time. Despite the promising results, these retrieval strategies suffer from two issues: \textbf{(1) Non-adaptive Retrieval Queries} refer to the fixed retrieval processes and query structures of heuristic mRAG methods. These inflexible retrieval strategies fail to adapt to evolving contexts or intermediate findings within a question, hindering the model from re-retrieving to further comprehend, verify, or rethink the question. For example, in Figure \ref{fig:intro_data}, question (a) asks, ``What is his (Cillian Murphy's) latest film?'' A fixed retrieval process returns multiple relevant films, but heuristic methods fail to construct further retrieval based on the retrieved content to distinguish between the sequence of different films. \textbf{(2) Overloaded Retrieval Queries} refer to heuristic retrieval methods refer to heuristic mRAG methods merely format a single query by concatenating textual descriptions of objects in images with input questions. A single query carries multiple retrieval aspects, leading to ambiguous retrieval and influx of superficially relevant knowledge yet not essential to the question solving. For example, in Figure \ref{fig:intro_data}, question (c) asks, ``Which one of them (two actors, Ling Jia, and Teng Shen) grossed more?'' Heuristic methods might generate a single query like ``Ling Jia, Teng Shen, Which one of them grossed more?'', which contains the intent to retrieve box office information for both actors. This mixed query conversely fails to provide precise knowledge for each individual aspect. Therefore, as shown in Figure \ref{fig:intro_method}, when faced with real-world questions requiring complex knowledge, current heuristic mRAG methods fail to provide sufficient and precise knowledge, due to their \textbf{\textit{rigidity issues}}.


Unfortunately, although several knowledge-seeking visual question answering (VQA) datasets \citep{chen2023can,schwenk2022okvqa} are widely utilized as mRAG benchmarks, they cannot adequately reflect the rigidity issues of heuristic mRAGs in acquiring complex knowledge. Since most questions in them merely require textual knowledge within two-hop, which can be readily obtained by heuristic mRAGs with a standard two-step retrieval process. As illustrated in the upper left of Figure \ref{fig:intro_data}, the most common type of question inquires about a certain property of an object.

To bridge the mRAG dataset gap, we first construct a challenging dataset, Dyn-VQA, comprising 1,452 dynamic questions that require complex multimodal knowledge retrieval for solution. \textbf{\textit{Dynamic questions}} are defined as questions that require the model to flexibly provide knowledge retrieval solutions, where the query, tool, and time of retrievals are all variable. These questions cannot be solved by a predefined retrieval process. Concretely, there types of dynamic questions are included in Dyn-VQA: \textbf{(1) Questions with rapidly changing answers.} Since the context knowledge of such question updates frequently, the retrieved content may be mixed with outdated and newer knowledge that is difficult to distinguish. This requires mARG methods to flexibly plan additional retrievals based on feedback from current retrieved content for further comprehension, rather than merely a one-time retrieval. \textbf{(2) Questions requiring multi-modal knowledge.} The knowledge necessary by Dyn-VQA spans various modalities. This demands that mRAG methods retrieve knowledge across diverse modalities with tailored retrieval APIs, differing from most VQA datasets limited in seeking textual knowledge with multimodal questions. \textbf{(3) Multi-hop questions.} Questions in Dyn-VQA necessitate varied reasoning hops for solution, which entails that mRAG methods conduct various retrieval steps. While existing VQA datasets primarily focus on two-hop question, i.e., identifying visual concepts via text and then answering single-hop textual question. 

We evaluated the performance of various mRAG methods combined with leading MLLMs on Dyn-VQA. Experiments reveal that existing heuristic mRAGs struggle to provide sufficient and precisely relevant knowledge for dynamic questions in Dyn-VQA, due to their rigid retrieval processes.

To address these issues, we further propose a self-adaptive planning agent for multimodal retrieval, \textbf{OmniSearch}\footnote{We aim for OmniSearch to achieve Omnipotent Multimodal Search, solving real-world issues in future.}. The underlying idea is to emulate the human behavior in question solution which dynamically decomposes complex multimodal questions into sub-question chains with retrieval action. At each step, OmniSearch flexibly adjusts the next action according to question-solving state and retrieved content, with diverse purposes such as deepening comprehension of retrieved content, modifying retrieval method for current sub-question, proposing the next sub-question, etc. It is noteworthy that OmniSearch can serve as a plug-and-play RAG module, cooperating with arbitrary MLLMs to augment their capability in addressing complex dynamic questions. Two different versions of OmniSearch are developed based on closed-source GPT-4V \citep{achiam2023gpt} and open-source Qwen-VL-Chat \citep{bai2023qwen}, respectively. As far as we know, OmniSearch is the first multimodal retrieval agent for VQA tasks with self-adaptive planning and scalable submodules.

In summary, our main contributions are fourfold:
\begin{itemize}[fullwidth,itemindent=1em]

 \item We reveal that current VQA-based mRAG benchmarks don't reflect real-world needs for dynamic knowledge retrieval and propose the Dyn-VQA dataset with three types of dynamic questions.

\item We benchmark various mRAG methods with leading MLLMs on Dyn-VQA, demonstrating their flaw in providing sufficient and relevant knowledge for dynamic questions.

\item We propose OmniSearch, a self-adaptive retrieval agent that plans each retrieval action in real-time according to question solution stage and current retrieval content.
\item Extensive experiments prove the effectiveness of our OmniSearch. Detailed analyses are conducted to provide direction for advancing mRAG.
\end{itemize}

\section{Related Works}
\label{sec:rw}

\subsection{Multimodal Retrieval Augmented Generation}
\label{sec:rw_mrag}
The mRAG methods \citep{ zhao2023retrieving, zhao2024retrieval, gao2023retrieval} aim to provide MLLMs \citep{lu2024deepseekvl,ye2024mplugowl,liu2024improved,du2022glm,chai2024mceval} with more comprehensive, accurate and up-to-date world knowledge from external sources~\citep{ wu2024tablebench,ji2024sevenllm, hou2024raw}. They have been empirically proven to be effective on various VQA datasets, which can be categorized into two classes based on the retrieval method.

One category employs visual encoding model for direct image representation, and then retrieves the knowledge from knowledge base with the highest feature similarity. For instance, KAT \citep{gui2022kat} and Revive \citep{lin2022revive} both use the image encoder of CLIP \citep{radford2021learning} for retrieval. The other category \citep{hu2023reveal,yang2022empirical,lin2024fine} first transforms the input image into textual representation utilizing off-the-shelf tools and then performs text retrieval. For example, RA-VQA \citep{lin2022retrieval} and RA-VQA-v2 \citep{lin2024fine} first use existing object detection and image captioning models to convert images to text, and than perform dense passage retrieval to fetch relevant text documents. Several studies \citep{yao2023react,xu2023rewoo} have preliminarily explored into agentic RAG, but their primary focus was on the text domain. \citet{chen-etal-2024-quantifying} proposed a causality-enhanced agent framework focus on unimodal biases in MLLMs, while it lacks plug-and-play capabilities.

The purpose of OmniSearch aligns with previous works to furnish pertinent and accurate knowledge for MLLMs, but diverging in three aspects: (1) OmniSearch plans multiple retrieval actions for the each question with diverse retrieval tools, supplementing the missing knowledge of each modality. (2) OmniSearch dynamically adjusts subsequent retrieval actions based on retrieved content, diverging from methods formulate query solely with input questions and images. (3) OmniSearch's retrieval scope extends to the entire Internet, offering intricate yet more comprehensive knowledge.

\section{Dyn-VQA Dataset}
In this section, we curated a novel dataset, Dyn-VQA, designed to evaluate the performance of existing mRAG methods in addressing questions requiring dynamic retrieval.

\subsection{Dyn-VQA Construction}
To guarantee the dataset quality, we explain the overall goal of the dataset to the annotators, who are all professional AI researchers. A straightforward construction strategy might directly request the annotators to write more visual questions after showing them the various types of VQA cases in Figure \ref{fig:intro_data}. However, in the preliminary annotation, we found that this overloaded single-step strategy is quite impractical. Annotators often found themselves in a dilemma of searching an image first, then laboriously devising a corresponding question while keeping various criteria, e.g., changing speed, and reasoning steps in their mind. Therefore, we optimize the strategy and divide them into three steps:

\noindent\textbf{Step 1. Textual Question Writing.} First, annotators are required to craft textual questions and categorize them using a three-dimensional schema based on answer update frequency (fast, slow, never), whether requiring external visual knowledge (yes, no), and reasoning steps ($\leq$ 2-hop, $>$ 2-hop). The annotation of fast or slow is determined by whether the updating occurs on yearly basis. Whether seeking visual knowledge beyond input images is also considered to separate questions emphasizing on different modalities. Meanwhile, multi-hop questions are delineated by a 2-hop cutoff, as previous VQA datasets typically focus on 2-hop. Annotators are prompted to compose questions incorporating newly emerged concepts from the past six months. The annotation difficulty is significantly reduced since visual information is not considered. Besides, English QA instances from FreshQA \citep{vu2023freshllms} are also included.

\noindent\textbf{Step 2. Multimodal Rewriting.}
The annotator transforms textual questions from the first step into multimodal ones by replacing visual concepts with co-references (e.g., "Kobe Bryant" to "this player") and pairing the revised question with a relevant image found on Google. Images sourced from commonly used pre-trained corpus such as Wikipedia and Baidu Encyclopedia are forbidden.

\noindent\textbf{Step 3. Chinese-English Translation.} 
This step aligns Chinese and English parts of the Dyn-VQA for side-by-side language comparison. Chinese and English VQA instances are translated into each other using Google Translate API, followed by manual checks and corrections to guarantee accuracy, especially for proper nouns. Instances that are intractable to translate or not applicable to Chinese/English contexts are filtered. Additionally, each question is annotated with the golden query, which simplifies the question by focusing solely on the last-hop question. This means that references to visual concepts and complex intermediate reasoning are omitted from the questions.

\subsection{Dyn-VQA Analysis}
\begin{table*}[th]
\centering
\vspace{-0.8cm}
\caption{Comparison of knowledge-seeking VQA datasets.}
\scalebox{0.55}{
\begin{tabular}{l|ccccccccc}
\toprule
\textbf{Dataset}          & \textbf{Knowledge Type} & \textbf{Ans. Change Freq.}      & \begin{tabular}[c]{@{}c@{}}\textbf{Reasoning}\\\textbf{Step}\end{tabular}     & \begin{tabular}[c]{@{}c@{}}\textbf{External}\\\textbf{Visual-Seek}\end{tabular} & \begin{tabular}[c]{@{}c@{}}\textbf{Human}\\\textbf{Annotation}\end{tabular} & \textbf{\# \{I, Q, A\}} & \begin{tabular}[c]{@{}c@{}}\textbf{Len. of}\\\textbf{Q/A}\end{tabular} & \textbf{Lang.} & \begin{tabular}[c]{@{}c@{}}\textbf{Const.}\\\textbf{Year}\end{tabular} \\ \midrule
VQAv2\citep{goyal2017making}            & common         & never change           & $\leq$ 2-hop  & \XSolidBrush                  & \Checkmark      & 614K              &  6.2/1.1           & en       & 2017        \\
OK-VQA \citep{marino2019ok}           & factoid        & never change           & 2-hop              & \XSolidBrush  & \Checkmark                     & 14K            &  8.1/1.3           & en       & 2019        \\ 
S3VQA \citep{jain2021select}            & factoid        & never change           & 2-hop              & \XSolidBrush & \XSolidBrush                      & 7.5K           &   12.7/2.8          & en       & 2021        \\ 
ViQuAE \citep{lerner2022viquae}          & fixed kb       & never change           & 2-hop              & \XSolidBrush & \XSolidBrush                       & 3.6K           &  10.9/2.4           & en       & 2022        \\ 
A-OKVQA \citep{schwenk2022okvqa}          & common/factoid & never change           & 2-hop              & \XSolidBrush  & \Checkmark                     & 24.9K          &  8.8/1.3           & en       & 2022        \\ 
InfoSeek \citep{chen2023can}        & fixed kb       & never change           & 2-hop              & \XSolidBrush  & \XSolidBrush                     & 1.35M          &  9.0/1.6           & en       & 2023        \\ \midrule
\textbf{Dyn-VQA} & real world     & fast/slow/never change & $>$ 2-hop & \Checkmark & \Checkmark                       & 1.5K           & 12.5/4.3            & zh/en    & 2024       \\ \bottomrule
\end{tabular}}
\label{tab:data_compare}
\end{table*}

\begin{wraptable}{r}{0.45\textwidth}
\centering
\vspace{-0.6cm}
\caption{Statistics of Dyn-VQA.}
\scalebox{0.63}{
\begin{tabular}{lc}
\toprule
\textbf{Statistic}                               &  \textbf{Number} \\ \midrule
Total Questions                                  &        1452                                \\
Domain                                           &            9                            \\
English questions                                & 715 (49.2\%)                                       \\
Chinese questions                                & 737 (50.8\%)                                  \\ \midrule
Questions with fast updating answers           & 385 (26.5\%)                                  \\
\qquad\&\& requiring $>$ 2-hop reasoning      &      112 (7.7\%)                                  \\
\qquad\&\& requiring external visual knowledge         &       178  (12.3\%)                               \\
Questions with slow updating answers           &  494 (34.0\%)                                      \\
Questions with never updating answers          &  573 (39.5\%)                                      \\ \midrule
Questions requiring $>$ 2-hop reasoning &   387 (26.7\%)                                     \\
\qquad\&\& requiring external visual knowledge          &                            237 (16.3\%)             \\
Questions requiring $\leq$ 2-hop reasoning &    1065 (73.3\%)                                    \\ \midrule
Questions requiring external visual knowledge     &   865 (59.6\%)                                     \\ 
Questions not requiring external visual knowledge     &   587 (40.4\%)                                     \\ \midrule
Average question length                          &  12.5                                 \\
Max question length                              &   60                                     \\
Average answer length                            &      4.3                                  \\
Max answer length                                &    73    \\ \bottomrule                               
\end{tabular}}
\vspace{-0.5cm}
\label{tab:statistic}
\end{wraptable}

\noindent\textbf{Dataset Statistics} Dyn-VQA is the first dataset specifically proposed for assessing the efficacy of mRAG systems. As shown in Table \ref{tab:statistic}, it contains $\sim$1.5K questions in 9 domains, covering 3 types of question requiring complex dynamic retrieval: questions with rapidly changing answers, questions requiring multi-modal knowledge, multi-hop questions. 

Comparative analysis between Dyn-VQA and other knowledge-seeking VQA datasets is also presented in Table \ref{tab:data_compare}. It is evident that while other datasets also emphasize the necessity of external knowledge in the question solving, their knowledge typically pertains to only one category. In contrast, each question in Dyn-VQA originates from the real world, encompassing a broader and more heterogeneous range of knowledge types, and featuring more open-ended answer styles. Reflecting in the length of questions and answers, Dyn-VQA exhibits the longest entries compared to other datasets. Furthermore, Dyn-VQA employs a more systematic question categorization schema, ensuring its challenging. Unlike other datasets, which are primarily constructed through templated and automated processes, Dyn-VQA is meticulously curated by humans and requires ongoing human input to maintain the dataset with dynamically updated answers. Consequently, while Dyn-VQA may not match other datasets in scale, it far surpasses them in terms of quality, difficulty, and the cost of each instance. \textbf{More details of Dyn-VQA can be found in Appendix.}

\noindent\textbf{Dataset Difficulty} Questions in Dyn-VQA require more complex external knowledge, whose retrieval process is not fixed. The inherent dynamism of Dyn-VQA naturally ensures its difficulty.
To illustrate more intuitively, we also conducted a quantitative comparison of human performance on different datasets. As shown in Table \ref{tab:difficult}, the questions in existing VQA datasets can typically be resolved within two reasoning steps, resulting in an average search count of less than 2 per question. Besides, image search with textual query is not performed at all in other datasets, since the question therein do not require additional visual knowledge except the textual description of the object in image. In terms of overall accuracy, humans achieved the lowest performance on Dyn-VQA, further demonstrating the challenges it presents.

\begin{table}[]
\centering
\caption{Human performances on different VQA datasets.}
\scalebox{0.7}{
\begin{tabular}{lcccc}
\toprule
\multirow{2}{*}{\textbf{Dataset}} & \multicolumn{3}{c}{\textbf{Search Count}}                                                                                                                                                  & \multirow{2}{*}{\textbf{Performance}} \\ \cmidrule{2-4}
                         & \textbf{Web Search} & \textbf{\begin{tabular}[c]{@{}c@{}}Image Search\\ with I. I.\end{tabular}} & \textbf{\begin{tabular}[c]{@{}c@{}}Image Search\\ with T. Q.\end{tabular}} &                              \\ \midrule
VQAv2                    & 0.05          & 0                             & 0                              & 74.31                        \\ 
A-OKVQA                  &   0.18         & 0.06                              & 0                               &  60.19                            \\ 
InfoSeek                 & 0.87           &  0.75                             & 0                               & 63.67                             \\ \midrule
Dyn-VQA                  & 1.57           & 0.89                              & 0.65                               &    55.12                         \\ \bottomrule
\end{tabular}}
\label{tab:difficult}
\vspace{-0.5cm}
\end{table}

\section{Retrieval Baselines and OmniSearch}
In this section, we establish several heuristic mRAG baselines and develop our OmniSearch to address dynamic questions that require complex retrieval. Retrieval tools in all methods are Google-based, including web search (textual web content retrieval with textual queries), image search with input images, and image search with textual query.
\subsection{Baselines}
\noindent\textbf{Single-hop heuristic mRAG baselines.} The first heuristic baseline is to \textbf{retrieve images with input images}, which provides similar images alongside descriptive captions. This method augments MLLMs with visual knowledge about the objects depicted in the input images. Similarly, the second heuristic baseline conducts \textbf{web search with input textual questions} and provides MLLMs with the top-k retrieved content as supplementary knowledge. It is acknowledged that these two methods may not furnish precise supportive knowledge, since them only leverage partial modality of the input question as search query. Nonetheless, these baselines are still established to explore the benefits from uni/cross-modality retrieval.

\noindent\textbf{Two-hop heuristic mRAG baselines.} Typically, existing mRAG methods can be generalized into two primary steps: first, converting the visual concepts in input images into text, and then combining the obtained text with input questions as search queries to retrieve related textual knowledge. For the first step, we employed the \textbf{caption of the top-1 image retrieved} by the input image and the \textbf{output of the image caption model} as textual representations of the input images, respectively. Subsequently, the web search API is utilized to extract relevant knowledge from the Internet. Compared to single-step mRAG methods, two-step mRAG methods can provide knowledge more precisely related to questions. However, they still encounter the limitation of a fixed retrieval process.

\noindent\textbf{Estimated Upper Bound.} We estimate the upper bound of mRAGs by the annotation of golden query, which simplifies the question to a last-hop question. Depending on whether the question requires external visual knowledge, either web search or image search API is called. We believe that in this simplified setting, the heuristic mRAG method can provide accurate information to MLLMs.

Descriptions of other baselines are available at Appendix.

\begin{figure*}[th]
\centering
\scalebox{0.9}{
\includegraphics[width=0.95\textwidth]{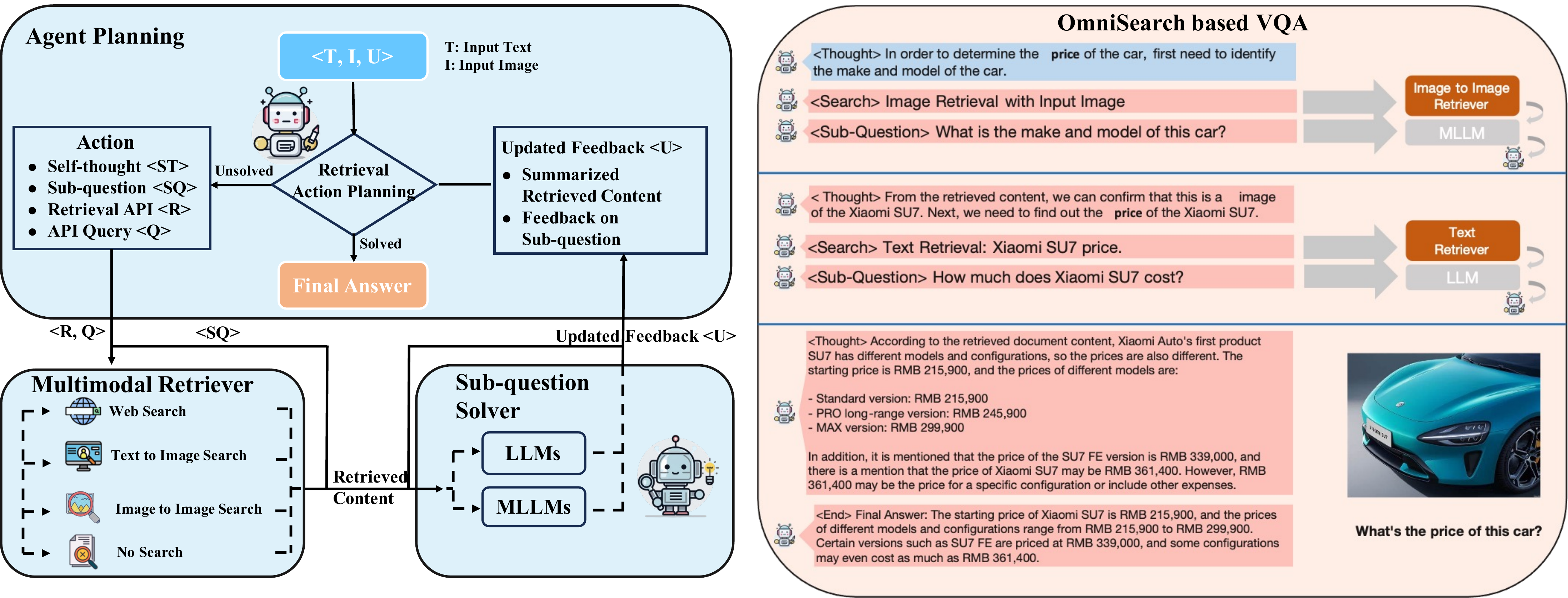}}
\caption{Left: overall framework of OmniSearch. Right: running example of OmniSearch.}
\label{fig3}
\vspace{-0.6cm}
\end{figure*}

\subsection{OmniSearch Framework}
Heuristic mRAG encounter the issues posed by rigid retrieval processes. A more flexible model that can plan the retrieval strategy tailored to the specific question and dynamically adjusts it throughout the process is desired. Therefore, we propose a self-routing framework called OmniSearch for retrieval planning. The underlying idea behind is to mimic the human of incrementally decomposing the complex question into a sequence of solution actions. As illustrated in Figure \ref{fig3}, overall framework consists of three module. The planning agent is the core module that formulates sub-questions and plans the subsequent retrieval action based on real-world feedback (i.e., retrieved content or solver output). Actual retrieval action is execute by the retriever. Then, sub-question solver generates feedback on sub-question based on the retrieval content and update it to the planner.

\textbf{Planning Agent}. Each planned action comprises four critical part: self-thought $\!<\! \!\text{ST}\!\! \! > $, sub-question  $<\!\!\!\text{SQ}\!\! \! > $, retrieval API $<\!\!\!\text{R}\!\! \! > $, and API query $<\!\!\!\text{Q}\!\! \! > $. In each step, planning agent comprehends the given question and the real-world feedback in self-thought, then carefully determines the follow-up sub-question to tackle. Meanwhile, different retrieval APIs with query are invoked, depending on the knowledge type required for the sub-question. In a manner akin to human cognitive processes during question-solving, planning agent autonomously generates various potential actions, including: posing additional question to clarify ambiguous or conflicting parts of the retrieved content; refining the retrieval query to acquire more supplementary knowledge; modifying the phrasing of sub-questions; verifying the response to the current sub-question; presenting the next sub-questions; concluding the final answer, etc.

\textbf{Retriever} executes actual retrieval operations. In our experiments, web search, image search with text and image retrieval with images are included. More retrieval tools can be considered in future.

\textbf{Sub-question Solver} summarizes the retrieved content and endeavors to address the sub-question accordingly. The feedback generated by the solver is then provided to the agent. Notably, the solver is compatible with arbitrary MLLMs or even the planning agent itself, i.e., directly returning the retrieved information to the planner. Depending on computational resources, MLLMs with larger or smaller sizes can be employed. Such pluggable and scalable feature is desired in industrial scenarios, which facilitates the flexible control of GPU cost.

The above process is fully automatic, with OmniSearch performing iterative actions until it believes it has retrieved sufficient knowledge to output a final answer. We trained two versions of OmniSearch based on different MLLMs: proprietary GPT-4V and open-source Qwen-VL-Chat. For GPT-4V, we employed prompt engineering to stimulate its dynamic planning and decision-making capabilities. To facilitate Qwen-VL-Chat's ability to plan and utilize retrieval APIs, we constructed a retrieval API training dataset, leveraging the GPT-4V synthetic data and the existing Infoseek dataset. We train the MLLM in a multi-round conversation mode. Additionally, general instruction data is also utilized to keep the general conversation capability of trained Agent.

The proposal of OmniSearch is inspired by Chain-of-Thought (CoT) \cite{wei2022chain}, but differs from it essentially. The fundamental distinction between OmniSearch, as a multimodal agent, and CoT is its ability to utilize tools, interact with the environment, and response to the environment \cite{zhang2023igniting}. In contrast, CoT methods primarily stimulate the model's inherent logical reasoning capabilities through prompts. The CoT approach is unable to decouple intermediate processes, therefore can not be integrated with retrieval tools.

\section{Experiments}
\subsection{Experimental Settings}
\paragraph{Backbone MLLMs for heuristic mRAGs.} We select several advanced MLLMs as backbone model to show their effectiveness equipped with heuristic mRAGs. \textbf{Qwen-VL-7B-Chat} is a large visual language model with strong visual and text recognition abilities proposed by \citet{Qwen-VL}. \textbf{GPT-4V} and \textbf{Qwen-VL-Max}\footnote{Our evaluation was conducted up to July 1, 2024.} are also selected as representatives of the closed-source models to show how the larger MLLMs will affect the results. Additionally, \textbf{Qwen-7B-Chat} is included in our experiments, which is a text-only LLM. We use this model to assess how multimodal RAG can solve the visual problem for textual LLM. 


\paragraph{Evaluation Metric.} The automated metric \textit{F1-Recall} is utilized as the evaluation metric. We calculate the ratio of common tokens between model-generated responses and ground truth. Specifically, we first segment the generated text and golden text into token lists using word segmentation tools\footnote{Jieba (\url{https://github.com/fxsjy/jieba}) for Chinese, and NLTK (\url{https://www.nltk.org/}) for English.}, then calculate the ratio of tokens generated by models belonging to the golden token list.

\subsection{Main Results}
\begin{table*}[tbh]
\vspace{-0.35cm}
\centering
\caption{Main results on Dyn-VQA.}
\scalebox{0.65}{
\begin{tabular}{lcccccccccc}
\toprule
\multirow{2}{*}{\textbf{Model}}   & \multicolumn{3}{c}{\textbf{Answer Update Frequency}}                     & \multicolumn{2}{c}{\textbf{Reasoning Steps}}         & \multicolumn{2}{c}{\textbf{Visual-Seeking}}          & \multicolumn{2}{c}{\textbf{Language}}                & \multirow{2}{*}{\textbf{all}} \\ \cmidrule{2-10}
                        & fast          & slow          & never          & $\leq$ 2-hop      & $>$ 2-hop     & no              & yes              & zh            & en           &                      \\ \midrule
\multicolumn{11}{l}{\textbf{\textit{Original (M)LLMs} }}                                   \\
Qwen-VL-Chat             &  13.69 & 14.20 & 17.27 & 15.56 & 14.49 & 15.50 & 15.13 & 16.92 & 13.58 & 15.28                  \\
Qwen-7B-Chat             & 5.63 & 7.86 & 15.97 & 10.48 & 10.43 & 11.05 & 10.08 & 10.48 & 10.46 & 10.47              \\
Qwen-VL-Max              &    15.11 & 30.44 & 39.51 & 30.22 & 29.21 & 31.10 & 29.18 & 22.81 & 37.32 & 29.96               \\
GPT-4V                   &   17.63 & 27.80 & 40.82 & 30.80 & 28.74 & 31.71 & 29.26 & 26.44 & 34.18 & 30.25                  \\ \midrule
\multicolumn{11}{l}{\textbf{+ \textit{Heuristic mRAG: Retrieving Images with Input Images} }}                                   \\
Qwen-VL-Chat            & 15.74 & 17.12 & 25.74 & 20.52 & 19.14 & 22.68 & 18.44 & 22.06 & 18.19 & 20.16                \\
Qwen-7B-Chat            & 10.97 & 15.04 & 25.91 & 18.76 & 16.86 & 24.18 & 14.23 & 16.80 & 19.75 & 18.25              \\
Qwen-VL-Max              &  24.04 & 28.99 & 45.49 & 34.22 & 34.08 & 41.54 & 29.19 & 31.07 & 37.39 & 34.19                 \\
GPT-4V                  & 20.18 & 33.21 & 50.00 & 35.94 & 35.65 & 42.63 & 31.33 & 30.90 & 40.32 & 35.87              \\ \midrule
\multicolumn{11}{l}{\textbf{+ \textit{Heuristic mRAG: Retrieving Web Pages with Input Questions} }}                                   \\
Qwen-VL-Chat            & 20.78 & 18.27 & 27.61 & 22.76 & 22.20 & 23.34 & 22.07 & 26.66 & 17.94 & 22.59               \\
Qwen-7B-Chat             & 14.65 & 15.47 & 24.98 & 19.14 & 18.66 & 19.99 & 18.34 & 17.93 & 20.12 & 19.01         \\
Qwen-VL-Max              & 26.71 & 27.37 & 35.84 & 30.65 & 30.22 & 31.14 & 30.13 & 30.27 & 30.82 & 30.54           \\
GPT-4V                   & 22.48 & 30.92 & 40.84 & 33.00 & 31.47 & 34.32 & 31.42 & 31.10 & 34.13 & 32.59                    \\ \midrule
\multicolumn{11}{l}{\textbf{+ \textit{Heuristic Two-Step mRAG: Retrieving Image First, then Retrieving Web Pages with Question appended to Retrieved Caption} }}                                   \\
Qwen-VL-Chat            & 19.17 & 20.02 & 28.54 & 23.33 & 22.68 & 23.84 & 22.69 & 24.11 & 22.17 & 23.16        \\
Qwen-7B-Chat            & 15.27 & 17.33 & 28.53 & 21.70 & 19.83 & 26.65 & 17.50 & 20.47 & 21.96 & 21.20         \\
Qwen-VL-Max              &   24.44 & 30.75 & 43.21 & 34.03 & 33.91 & 38.26 & 31.10 & 32.04 & 36.01 & 33.99                \\
GPT-4V                 & 20.37 & 33.98 & 48.46 & 36.12 & 36.04 & 40.19 & 33.32 & 32.99 & 39.30 & 36.10         \\ \midrule
\multicolumn{11}{l}{\textbf{+ \textit{Heuristic Two-Step mRAG: Image Caption First, then Retrieving Web Pages with Question appended to Caption} }}                                   \\
Qwen-VL-Chat            & 22.05 & 25.87 & 31.84 & 27.58 & 26.21 & 27.44 & 27.06 & 28.81 & 25.57 & 27.21             \\
Qwen-7B-Chat            & 14.65 & 21.16 & 28.66 & 22.89 & 21.02 & 23.64 & 21.55 & 16.57 & 28.39 & 22.39              \\
Qwen-VL-Max              &  24.27 & 32.93 & 44.03 & 35.04 & 34.97 & 35.16 & 34.92 & 31.10 & 39.05 & 35.02                    \\
GPT-4V                  & 24.90 & 36.74 & 45.76 & 37.23 & 36.94 & 37.82 & 36.70 & 31.65 & 42.81 & 37.15            \\ \midrule
\multicolumn{11}{l}{\textbf{\textit{Generative Search Engine} }}                                   \\
Bing Chat                & 27.71 & 32.77 & 32.99 & 31.67 & 30.80 & 35.44 & 28.64 & 29.62 & 32.74 & 31.44                   \\
Perplexity AI            &   29.62 & 34.69 & 34.88 & 33.67 & 32.81 & 37.46 & 30.67 & 31.59 & 34.80 & 33.51            \\ 
Gemini                  & 36.17 & 32.86 & 42.84 & 38.75 & 34.78 & 46.39 & 31.82 & 35.77 & 39.69 & 37.69  \\ \midrule
\multicolumn{11}{l}{\textbf{\textit{Ours} }}   \\
OmniSearch (Qwen-VL-Chat) &  35.16 & 40.89 & 45.52 & 41.34 & 40.81 & 42.56 & 40.28 & 39.22 & 43.23 & 41.20             \\
OmniSearch (GPT-4V)       & \textbf{44.04} & \textbf{49.58} & \textbf{54.45} & \textbf{50.38} & \textbf{49.06} & \textbf{50.49} & \textbf{49.73} & \textbf{46.96} & \textbf{53.21} & \textbf{50.03}                     \\ \midrule
\multicolumn{11}{l}{\textbf{\textit{Estimated Upper Bound: + Retrieving Web Pages with Gloden Query} }}   \\
Qwen-VL-Chat            & 37.46 & 46.52 & 52.18 & 46.73 & 45.28 & 47.94 & 45.27 & 43.88 & 48.90 & 46.35           \\
Qwen-7B-Chat            & 39.69 & 47.27 & 57.76 & 49.53 & 49.02 & 50.94 & 48.36 & 46.02 & 52.88 & 49.40     \\
Qwen-VL-Max              & 42.19 & 53.01 & 56.60 & 51.91 & 50.58 & 52.97 & 50.60 & 49.83 & 53.33 & 51.56                  \\
GPT-4V                  & 45.59 & 54.23 & 60.78 & 55.15 & 52.81 & 54.53 & 54.51 & 51.08 & 58.07 & 54.52             \\ \midrule
Human Performance                & 51.63 & 60.02 & 53.19 & 54.12 & 58.31 & 57.86 & 53.20 & 51.96 & 58.36 & 55.12 \\ \bottomrule              
\end{tabular}}
\label{tab:main}
\vspace{-0.35cm}
\end{table*}
The performance of various MLLMs with different mRAG methods are shown in Table \ref{tab:main}, from which we can find that: 

(1) Our OmniSearch (GPT-4V) significantly outperforms other models, encompassing both state-of-the-art MLLMs with heuristic mRAGs and commercial generative search engines. Even Qwen-VL-Chat-based OmniSearch surpasses the considerably larger GPT-4V equipped with two-step heuristic mRAG. We attribute this to two aspects: on the one hand, the OmniSearch decomposes a complex question into a sequence of sub-questions, reducing the retrieval burden in a single step. On the other hand, it rethinks the retrieved content and sub-questions to ensure the accuracy of the sub-answers, mitigating the risk of error propagation.

(2) Regarding overall performance, the OmniSearch closely parallel human and GPT-4V enhanced with content retrieved via gold query, highlighting its superior abilities. Nevertheless, a significant gap remains between the OmniSearch and human performance on questions belong to the three most challenging subcategories (fast-changing, $>$2-hop, requiring external visual knowledge), which indicates substantial room for improvement in agent-based mRAG for real-world questions. How to generate more human-like search logic is a promising direction for future research.

(3) Despite achieving more than 50\% performance, Dyn-VQA remains a formidable challenge for both AI systems and humans. It is observed that for questions necessitating multi-step retrieval or additional visual knowledge, all models consistently underperform compared to other questions within the same classification schema. Especially for questions with different answer update frequencies, the variance in model performance is high. We can conclude that questions requiring rapidly changing knowledge pose the most intractable challenge, as such knowledge cannot be internalized by MLLMs.

(4) For two-step heuristic mRAGs, leveraging image caption model to transform visual concepts brings more gain to the original MLLMs, which provides a more detailed image description for the next retrieval step. However, this advantage reverses for questions that do not require additional visual knowledge, primarily because the majority of them are 2-hop (74\%) and do not demand visual knowledge beyond the concepts presented in the image itself. Supplementary information from the image caption model does not substantially benefit the model.

(5) Commercial generative search engines generally perform poorly on Dyn-VQA. Even the best-performing engine, Gemini, only matching the performance of GPT-4V with two-step mRAG. Further case analysis reveals these generative search engines lack essential grounding capabilities: they fail to associate ``it'' in the question with objects in the image, nor can integrate multimodal information effectively. This suggests that questions in Dyn-VQA represent the real demand in industrial scenarios.

(6) Comparing Qwen-7B-Chat and Qwen-VL-Chat, we observe that the performance gap between the models is reduced once equipped with mRAG. This phenomenon indicates that mRAG can assist pure text LLMs in addressing multi-modal issues.

Due to space constraints, more analysis experiments are placed \textbf{in the appendices in the Supplementary Material, and they are highly recommended to the reader}.
\begin{table*}[th]
\centering
\vspace{-0.5cm}
\caption{Experiments on OmniSearch paired with different MLLMs as sub-question solvers. OmniSearch (G) and OmniSearch (Q) refer to OmniSearch implementations based on GPT-4V and Qwen-VL-Chat, respectively.}
\scalebox{0.65}{
\begin{tabular}{llcccccccccc}
\toprule
\multirow{2}{*}{\textbf{Planning Model}} & \multirow{2}{*}{\textbf{Sub-question Solver}} & \multicolumn{3}{c}{\textbf{Answer Update Frequency}} & \multicolumn{2}{c}{\textbf{Reasoning Steps}} & \multicolumn{2}{c}{\textbf{Visual-Seeking}} & \multicolumn{2}{c}{\textbf{Language}} & \multirow{2}{*}{\textbf{all}} \\ \cmidrule{3-11}
                                &                                      & fast          & slow          & never          & $\leq$ 2-hop      & $>$ 2-hop          & no              & yes              & zh            & en           &                      \\ \midrule
OmniSearch (Q)        & OmniSearch (Q)             &   35.16 & 40.89 & 45.52 & 41.34 & 40.81 & 42.56 & 40.28 & 39.22 & 43.23 & 41.20                    \\
OmniSearch (Q)        & GPT-4V                               & 37.14 & 42.82 & 47.48 & 43.29 & 42.78 & 44.46 & 42.26 & 41.21 & 45.15 & 43.15    \\
OmniSearch (Q)        & GPT-4V + GPT-4                     &   38.98 & 44.52 & 49.18 & 45.03 & 44.52 & 46.20 & 44.00 & 42.97 & 46.87 & 44.89     \\
OmniSearch (Q)        & Qwen-VL-Chat                         &  34.10 & 39.88 & 44.50 & 40.32 & 39.77 & 41.53 & 39.25 & 38.18 & 42.22 & 40.17     \\ \midrule
OmniSearch (G)              & OmniSearch (G)                   &  44.04 & 49.58 & 54.45 & 50.38 & 49.06 & 50.49 & 49.73 & 46.96 & 53.21 & 50.03            \\
OmniSearch (G)              & Qwen-VL-Chat                         &  38.65 & 44.68 & 52.25 & 46.56 & 44.72 & 49.40 & 43.80 & 41.63 & 50.64 & 46.07             \\ \bottomrule
\end{tabular}}
\vspace{-0.5cm}
\label{tab:exp_sub_solver}
\end{table*}

\subsection{Analysis Experiments on OmniSearch}
 In this section, we conduct extensive analysis experiments to answer the following questions on our OmniSearch:
 
\noindent\textbf{How different models as sub-question solvers affect overall performance?} As shown in Table 
\ref{tab:exp_sub_solver}, several observations can be made regarding the performance of OmniSearch when paired with different MLLMs as sub-question solvers:

(1) In the case of the Qwen-VL-Chat Based OmniSearch, employing the larger model GPT-4V as the sub-question solver significantly enhances the performance of the OmniSearch, indicating the continued validity of scaling laws for sub-question solver. Meanwhile, substituting the sub-question solver of the GPT-4V Based OmniSearch with the smaller Qwen-VL-Chat leads to a predictable decrease in model performance. Nonetheless, it still outperforms Qwen-VL-Chat with two-step heuristics mRAG from Table \ref{tab:main}.

(2) we also explored a more complex invocation strategy for the sub-question solving model: leveraging GPT-4V for sub-questions entailing multimodal contexts, and employing GPT-4 for those involving purely textual contexts, which is considered to be more capable on text-only questions compared to GPT-4V. This approach contributes further to performance enhancement. More refined invocation strategies is also worthy to be explored, such as having the sub-question solver output bounding boxes for certain objects in the image to guide more precise retrieval. We leave this topic in the future work.

(3) To assess whether the OmniSearch with retrieval path planning learning has been impaired in its question solving ability, we replaced the sub-question solver of the Qwen-VL-Chat Based OmniSearch with the original Qwen-VL-Chat. Comparison of rows 1 and 4 in Table \ref{tab:exp_sub_solver} reveals that employing the OmniSearch as the sub-question solver instead improves question-solving ability. This enhancement demonstrate that the learning of retrieval path planning also involves the ability to understand and reason about retrieval knowledge, potentially enhancing the model's problem solving ability and yielding cross-task gains.

\begin{table*}[]
\centering
\caption{Comparison of token costs and expenses for different models.}
\scalebox{0.7}{
\begin{tabular}{llllcc}
\toprule
\textbf{Planning Model }         & \textbf{Sub-question Solver} & \textbf{\# Input Tokens} & \textbf{\# Output Tokens} & \textbf{Expenses ($\times10^{-3}\$$)} & \textbf{Performance} \\ \midrule
Two-Step mRAG & GPT-4V                & 1454.0 (G)           & 132.5 (G)            & 18.5  & 37.15       \\
Two-Step mRAG & Qwen-VL-Chat          & 749.9 (Q)           & 28.6 (Q)               & 0.2     & 27.21            \\ \midrule
OmniSearch (G)        & OmniSearch (G)      & 3028.5 (G)                & 476.9 (G)                 & 44.6     & 50.03            \\ 
OmniSearch (G)        & Qwen-VL-Chat          & 1217.2 (G) + 2073.4 (Q)                & 386.0 (G) + 124.8 (Q)                 &     24.4  & 46.07            \\
OmniSearch (Q)        & OmniSearch (Q)      & 9578.3 (Q)               & 572.5 (Q)                 & 3.2     & 41.20            \\
OmniSearch (Q)        & GPT-4V                &  2371.5 (G) + 992.2 (Q)              & 281.0 (G) + 551.4 (Q)                &  32.8    & 43.15            \\  \bottomrule
\end{tabular}}
\label{tab:token_cost}
\end{table*}

\noindent\textbf{How different models as sub-question solvers affect token and expenses?} 
In Table \ref{tab:token_cost}, we further examine the token costs and actual expenses\footnote{We refer https://azure.microsoft.com/zh-cn/pricing/details/cognitive-services/openai-service/ for GPT-4V price.} brought by different sub-question solvers. Although more costly, the enhancement provided by the OmniSearch relative to the heuristic mRAG is considerably substantial. The correlation between the performance of the OmniSearch and the actual expenses is proportional, yet not linear. When substituting GPT-4V with Qwen-VL-Chat as the sub-question solver (rows 3 and 4), the absolute performance declines by under 4 points, approximately 7.9\%, while the expenditure is nearly halved, demonstrating the excellent scalability of OmniSearch. The results also indicate that sub-question reasoning is not the bottleneck of current methods, rather the retrieval strategy of complex questions presents a more urgent challenge. This conclusion is further supported by comparing the enhancements achieved by substituting the planning model (row 6 to 5) and sub-question solver (row 6 to 7) of the OmniSearch (Q) with GPT-4V. The benefits realized by the former are more pronounced. Consequently, when computational resources are constrained, priority should be given to ensuring that retrieval planning model can utilize a larger model as backbone.

\section{Analysis Experiments on Dyn-VQA Dataset}
\subsection{Performance Comparison on Different Domains}
\label{sec:domain_perf}
Figure \ref{fig:radar} illustrates the performance of Qwen-VL-Chat and GPT-4V equipped with different mRAG methods across various domains. We can intuitively observe that each mRAG method enhances the efficacy of the original model. The coverage of both original models is notably broadened by the mRAG methods, particularly for the smaller Qwen-VL-Chat. However, in several domains, such as transportation, the Qwen-VL-Chat-based OmniSearch instead exhibits superior performance compared to the GPT-4V-based one. Further analysis reveals that this phenomenon is primarily attributed to the long-tail property of transportation domain, which contains only 10 VQA instances, with the majority comprising 2-hop questions or questions that do not involve changing knowledge. In these cases, GPT-4V-based OmniSearch tends to over-retrieve, e.g., it has retrieved the necessary information but over-cautiously continues to gather additional information to validate the answer, resulting in the correct answer being obscured within a large volume of retrieved information. This underscores the need for ongoing enhancements to the robustness of our OmniSearch.

\subsection{Prediction Overlap}
\begin{figure}[t]
\centering
\begin{minipage}[t]{0.45\textwidth} 
\centering
\scalebox{0.9}{
\includegraphics[width=0.9\textwidth]{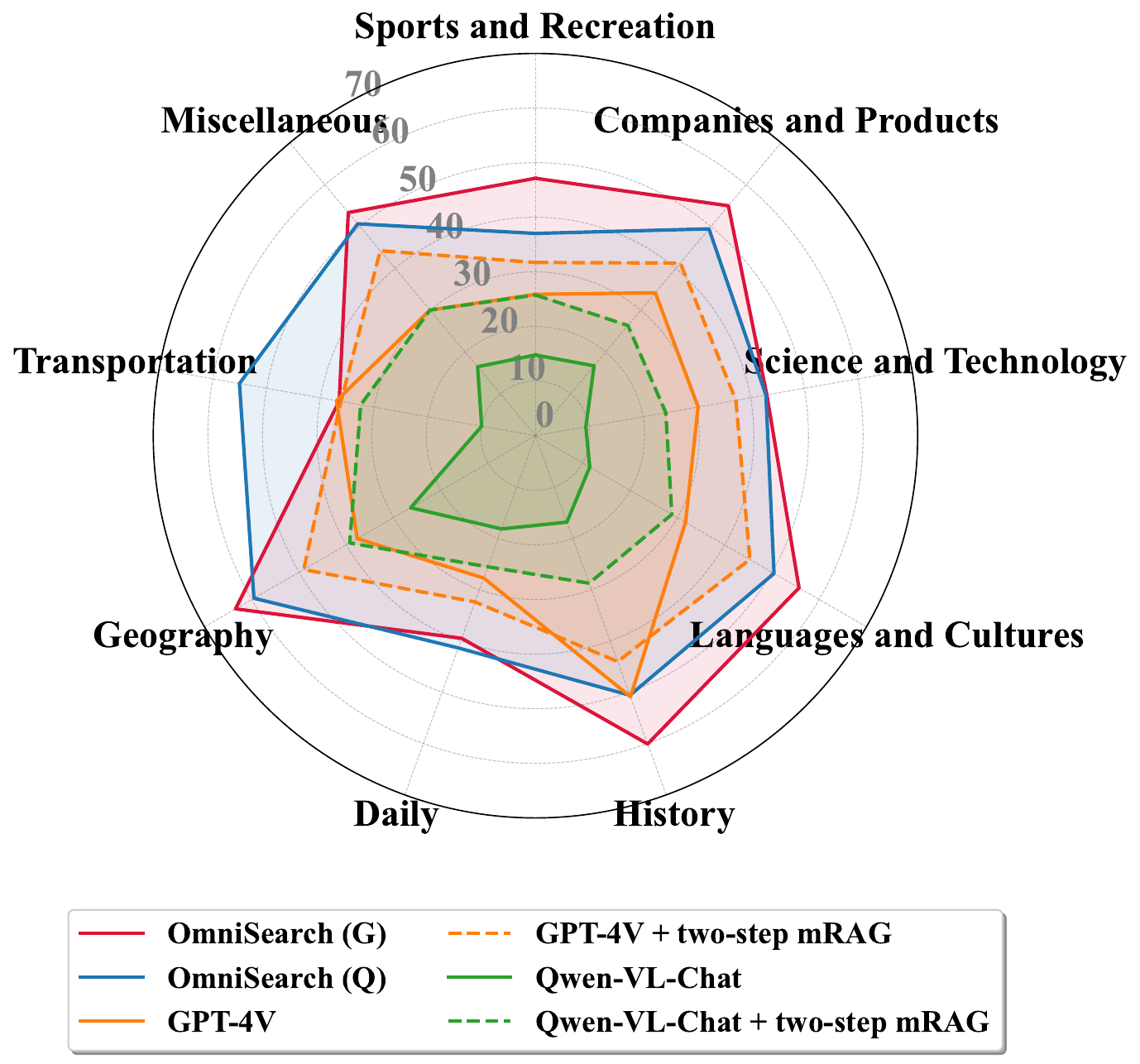}}
\caption{Model performance of different domains.}
\vspace{-0.2cm}
\label{fig:radar}
\vspace{-0.5cm}
\end{minipage}%
\hspace{0.1cm}
\begin{minipage}[t]{0.5\textwidth} 
\centering
\scalebox{0.9}{
\includegraphics[width=0.9\textwidth]{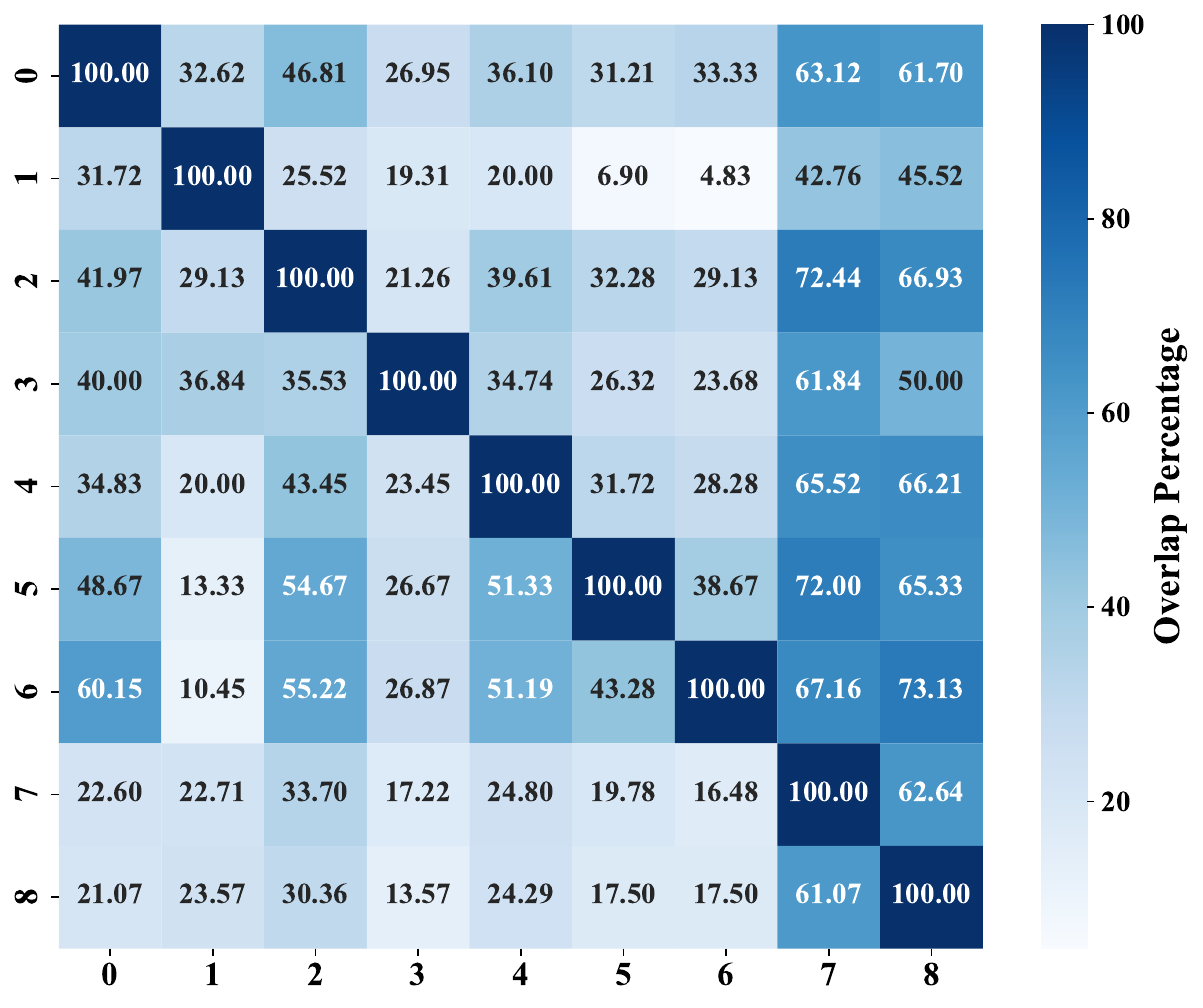}}
\caption{Pairwise overlap between correctly answered questions of different models. 0-8: Qwen-VL-Chat, Qwen-7B-Chat, Deepseek-VL-7B-Chat, VisualGLM-6B, Llava-V1.6-Mistral-7B, mPLUG-Owl2.1, InstructBLIP-Vicuna-7B, Qwen-VL-Max, GPT-4V.}
\label{fig:hp-heatmap}
\end{minipage}

\end{figure}

In this subsection, we investigate the overlap of questions correctly answered by different models. Firstly, we observed that no questions in the Dyn-VQA were correctly answered by all models, and 31\% of the questions did not receive a correct prediction from any model. Figure \ref{fig:hp-heatmap} illustrates the degree of pairwise overlap in correctly answering questions on Dyn-VQA. Each row indicates the proportion of questions correctly answered by the corresponding model that were also correctly answered by other models. Overall, the two highest-performing models, Qwen-VL-Max and GPT-4V, exhibited relatively high overlap,  but still hover around 60\%. Furthermore, looking at heat blocks (6, 8), we find that even for InstructBLIP-Vicuna-7B, which demonstrated the weakest performance (12.33 overall F1 recall, as detailed in Table 11 in the Appendix), 26.87\% (100 - 73.13) of the questions it successfully answered could not be correctly answered by the best-performing GPT-4V. This indicates substantial differences in model behavior and shows that although some models generally outperform others, their superiority is not attributable to correctly answering the ``hard'' questions while consistently getting the ``easy'' ones right. The varied challenges presented by Dyn-VQA affect models differently, highlighting ensemble-based and self-consist-based approaches as promising directions for future research. 

\section{Conclusion}
In this paper, we study the multimodal retrieval augmented generation (mRAG). We argue that existing heuristic mRAGs typically predefined fixed retrieval processes, which causes two issues: (1) Non-adaptive Retrieval Queries. (2) Overloaded Retrieval Queries. However, these rigidity issues cannot be adequately reflected by current knowledge-seeking visual question answering (VQA) datasets. Therefore, we first construct Dyn-VQA dataset, consisting of three types of ``dynamic'' questions, which require complex knowledge retrieval strategies variable in query, tool, and time. Furthermore, OmniSearch is proposed as the first self-adaptive planning agent for multimodal retrieval. Extensive experiments prove the effectiveness of our OmniSearch, also highlight the challenges posed by Dyn-VQA.

\section*{Acknowledgements}
This research project is supported by National Natural Science Foundation of China (Grant No.62276154), Research Center for ComputerNetwork (Shenzhen) Ministry of Education, the Natural Science Foundation of Guangdong Province (Grant No.2023A1515012914 and 440300241033100801770), Basic Research Fund of Shenzhen City (Grant No.JCYJ20210324120012033, JCYJ20240813112009013 and GJHZ20240218113603006), the Major Key Projectof PCL for Experiments and Applications (PCL2022A05 and PCL2023A09), NSF under Grants III-2106758, and POSE-2346158.

\bibliography{iclr2025_conference}

\begin{thebibliography}{52}
\providecommand{\natexlab}[1]{#1}
\providecommand{\url}[1]{\texttt{#1}}
\expandafter\ifx\csname urlstyle\endcsname\relax
  \providecommand{\doi}[1]{doi: #1}\else
  \providecommand{\doi}{doi: \begingroup \urlstyle{rm}\Url}\fi

\bibitem[Achiam et~al.(2023)Achiam, Adler, Agarwal, Ahmad, Akkaya, Aleman, Almeida, Altenschmidt, Altman, Anadkat, et~al.]{achiam2023gpt}
Josh Achiam, Steven Adler, Sandhini Agarwal, Lama Ahmad, Ilge Akkaya, Florencia~Leoni Aleman, Diogo Almeida, Janko Altenschmidt, Sam Altman, Shyamal Anadkat, et~al.
\newblock Gpt-4 technical report.
\newblock \emph{arXiv preprint arXiv:2303.08774}, 2023.

\bibitem[Bai et~al.(2023{\natexlab{a}})Bai, Bai, Chu, Cui, Dang, Deng, Fan, Ge, Han, Huang, et~al.]{bai2023qwen}
Jinze Bai, Shuai Bai, Yunfei Chu, Zeyu Cui, Kai Dang, Xiaodong Deng, Yang Fan, Wenbin Ge, Yu~Han, Fei Huang, et~al.
\newblock Qwen technical report.
\newblock \emph{arXiv preprint arXiv:2309.16609}, 2023{\natexlab{a}}.

\bibitem[Bai et~al.(2023{\natexlab{b}})Bai, Bai, Yang, Wang, Tan, Wang, Lin, Zhou, and Zhou]{Qwen-VL}
Jinze Bai, Shuai Bai, Shusheng Yang, Shijie Wang, Sinan Tan, Peng Wang, Junyang Lin, Chang Zhou, and Jingren Zhou.
\newblock Qwen-vl: A versatile vision-language model for understanding, localization, text reading, and beyond.
\newblock \emph{arXiv preprint arXiv:2308.12966}, 2023{\natexlab{b}}.

\bibitem[Bai et~al.(2024)Bai, Wang, Xiao, He, Han, Zhang, and Shou]{bai2024hallucination}
Zechen Bai, Pichao Wang, Tianjun Xiao, Tong He, Zongbo Han, Zheng Zhang, and Mike~Zheng Shou.
\newblock Hallucination of multimodal large language models: A survey.
\newblock \emph{arXiv preprint arXiv:2404.18930}, 2024.

\bibitem[Brown et~al.(2020)Brown, Mann, Ryder, Subbiah, Kaplan, Dhariwal, Neelakantan, Shyam, Sastry, Askell, et~al.]{brown2020language}
Tom Brown, Benjamin Mann, Nick Ryder, Melanie Subbiah, Jared~D Kaplan, Prafulla Dhariwal, Arvind Neelakantan, Pranav Shyam, Girish Sastry, Amanda Askell, et~al.
\newblock Language models are few-shot learners.
\newblock \emph{Advances in neural information processing systems}, 33:\penalty0 1877--1901, 2020.

\bibitem[Chai et~al.(2024{\natexlab{a}})Chai, Liu, Yang, Yin, Jin, Liu, Sun, Zhang, Ren, Guo, et~al.]{chai2024mceval}
Linzheng Chai, Shukai Liu, Jian Yang, Yuwei Yin, Ke~Jin, Jiaheng Liu, Tao Sun, Ge~Zhang, Changyu Ren, Hongcheng Guo, et~al.
\newblock Mceval: Massively multilingual code evaluation.
\newblock \emph{arXiv preprint arXiv:2406.07436}, 2024{\natexlab{a}}.

\bibitem[Chai et~al.(2024{\natexlab{b}})Chai, Yang, Sun, Guo, Liu, Wang, Liang, Bai, Li, Peng, and Li]{xcot}
Linzheng Chai, Jian Yang, Tao Sun, Hongcheng Guo, Jiaheng Liu, Bing Wang, Xinnian Liang, Jiaqi Bai, Tongliang Li, Qiyao Peng, and Zhoujun Li.
\newblock xcot: Cross-lingual instruction tuning for cross-lingual chain-of-thought reasoning.
\newblock \emph{arXiv preprint arXiv:2401.07037}, abs/2401.07037, 2024{\natexlab{b}}.
\newblock \doi{10.48550/ARXIV.2401.07037}.
\newblock URL \url{https://doi.org/10.48550/arXiv.2401.07037}.

\bibitem[Chen et~al.(2024{\natexlab{a}})Chen, Cao, Zhang, and Lu]{chen-etal-2024-quantifying}
Meiqi Chen, Yixin Cao, Yan Zhang, and Chaochao Lu.
\newblock Quantifying and mitigating unimodal biases in multimodal large language models: A causal perspective.
\newblock In Yaser Al-Onaizan, Mohit Bansal, and Yun-Nung Chen (eds.), \emph{Findings of the Association for Computational Linguistics: EMNLP 2024}, pp.\  16449--16469, Miami, Florida, USA, November 2024{\natexlab{a}}. Association for Computational Linguistics.
\newblock \doi{10.18653/v1/2024.findings-emnlp.960}.
\newblock URL \url{https://aclanthology.org/2024.findings-emnlp.960}.

\bibitem[Chen et~al.(2023)Chen, Hu, Luan, Sun, Changpinyo, Ritter, and Chang]{chen2023can}
Yang Chen, Hexiang Hu, Yi~Luan, Haitian Sun, Soravit Changpinyo, Alan Ritter, and Ming-Wei Chang.
\newblock Can pre-trained vision and language models answer visual information-seeking questions?
\newblock \emph{arXiv preprint arXiv:2302.11713}, 2023.

\bibitem[Chen et~al.(2024{\natexlab{b}})Chen, Sikka, Cogswell, Ji, and Divakaran]{chen2024measuring}
Yangyi Chen, Karan Sikka, Michael Cogswell, Heng Ji, and Ajay Divakaran.
\newblock Measuring and improving chain-of-thought reasoning in vision-language models.
\newblock In \emph{Proceedings of the 2024 Conference of the North American Chapter of the Association for Computational Linguistics: Human Language Technologies (Volume 1: Long Papers)}, pp.\  192--210, 2024{\natexlab{b}}.

\bibitem[Dai et~al.(2023)Dai, Li, Li, Tiong, Zhao, Wang, Li, Fung, and Hoi]{instructblip}
Wenliang Dai, Junnan Li, Dongxu Li, Anthony Meng~Huat Tiong, Junqi Zhao, Weisheng Wang, Boyang Li, Pascale Fung, and Steven Hoi.
\newblock Instructblip: Towards general-purpose vision-language models with instruction tuning, 2023.

\bibitem[Du et~al.(2022)Du, Qian, Liu, Ding, Qiu, Yang, and Tang]{du2022glm}
Zhengxiao Du, Yujie Qian, Xiao Liu, Ming Ding, Jiezhong Qiu, Zhilin Yang, and Jie Tang.
\newblock Glm: General language model pretraining with autoregressive blank infilling.
\newblock In \emph{Proceedings of the 60th Annual Meeting of the Association for Computational Linguistics (Volume 1: Long Papers)}, pp.\  320--335, 2022.

\bibitem[Fleiss(1971)]{fleiss1971measuring}
Joseph~L Fleiss.
\newblock Measuring nominal scale agreement among many raters.
\newblock \emph{Psychological bulletin}, 76\penalty0 (5):\penalty0 378, 1971.

\bibitem[Gao et~al.(2023)Gao, Xiong, Gao, Jia, Pan, Bi, Dai, Sun, and Wang]{gao2023retrieval}
Yunfan Gao, Yun Xiong, Xinyu Gao, Kangxiang Jia, Jinliu Pan, Yuxi Bi, Yi~Dai, Jiawei Sun, and Haofen Wang.
\newblock Retrieval-augmented generation for large language models: A survey.
\newblock \emph{arXiv preprint arXiv:2312.10997}, 2023.

\bibitem[Goyal et~al.(2017)Goyal, Khot, Summers-Stay, Batra, and Parikh]{goyal2017making}
Yash Goyal, Tejas Khot, Douglas Summers-Stay, Dhruv Batra, and Devi Parikh.
\newblock Making the v in vqa matter: Elevating the role of image understanding in visual question answering.
\newblock In \emph{Proceedings of the IEEE conference on computer vision and pattern recognition}, pp.\  6904--6913, 2017.

\bibitem[Gui et~al.(2022)Gui, Wang, Huang, Hauptmann, Bisk, and Gao]{gui2022kat}
Liangke Gui, Borui Wang, Qiuyuan Huang, Alexander~G Hauptmann, Yonatan Bisk, and Jianfeng Gao.
\newblock Kat: A knowledge augmented transformer for vision-and-language.
\newblock In \emph{Proceedings of the 2022 Conference of the North American Chapter of the Association for Computational Linguistics: Human Language Technologies}, pp.\  956--968, 2022.

\bibitem[Hou et~al.(2024)Hou, Li, Yang, Li, Chai, Wu, Ji, Li, Nie, Dun, et~al.]{hou2024raw}
Xia Hou, Qifeng Li, Jian Yang, Tongliang Li, Linzheng Chai, Xianjie Wu, Hangyuan Ji, Zhoujun Li, Jixuan Nie, Jingbo Dun, et~al.
\newblock Raw text is all you need: Knowledge-intensive multi-turn instruction tuning for large language model.
\newblock \emph{arXiv preprint arXiv:2407.03040}, 2024.

\bibitem[Hu et~al.(2023)Hu, Iscen, Sun, Wang, Chang, Sun, Schmid, Ross, and Fathi]{hu2023reveal}
Ziniu Hu, Ahmet Iscen, Chen Sun, Zirui Wang, Kai-Wei Chang, Yizhou Sun, Cordelia Schmid, David~A Ross, and Alireza Fathi.
\newblock Reveal: Retrieval-augmented visual-language pre-training with multi-source multimodal knowledge memory.
\newblock In \emph{Proceedings of the IEEE/CVF conference on computer vision and pattern recognition}, pp.\  23369--23379, 2023.

\bibitem[Jain et~al.(2021)Jain, Kothyari, Kumar, Jyothi, Ramakrishnan, and Chakrabarti]{jain2021select}
Aman Jain, Mayank Kothyari, Vishwajeet Kumar, Preethi Jyothi, Ganesh Ramakrishnan, and Soumen Chakrabarti.
\newblock Select, substitute, search: A new benchmark for knowledge-augmented visual question answering.
\newblock In \emph{Proceedings of the 44th International ACM SIGIR Conference on Research and Development in Information Retrieval}, pp.\  2491--2498, 2021.

\bibitem[Jain et~al.(2024)Jain, Yang, and Shi]{jain2024vcoder}
Jitesh Jain, Jianwei Yang, and Humphrey Shi.
\newblock Vcoder: Versatile vision encoders for multimodal large language models.
\newblock In \emph{Proceedings of the IEEE/CVF Conference on Computer Vision and Pattern Recognition}, pp.\  27992--28002, 2024.

\bibitem[Ji et~al.(2024)Ji, Yang, Chai, Wei, Yang, Duan, Wang, Sun, Guo, Li, et~al.]{ji2024sevenllm}
Hangyuan Ji, Jian Yang, Linzheng Chai, Chaoren Wei, Liqun Yang, Yunlong Duan, Yunli Wang, Tianzhen Sun, Hongcheng Guo, Tongliang Li, et~al.
\newblock Sevenllm: Benchmarking, eliciting, and enhancing abilities of large language models in cyber threat intelligence.
\newblock \emph{arXiv preprint arXiv:2405.03446}, 2024.

\bibitem[Jiang et~al.(2023)Jiang, Sablayrolles, Mensch, Bamford, Chaplot, de~las Casas, Bressand, Lengyel, Lample, Saulnier, Lavaud, Lachaux, Stock, Scao, Lavril, Wang, Lacroix, and Sayed]{jiang2023mistral}
Albert~Q. Jiang, Alexandre Sablayrolles, Arthur Mensch, Chris Bamford, Devendra~Singh Chaplot, Diego de~las Casas, Florian Bressand, Gianna Lengyel, Guillaume Lample, Lucile Saulnier, Lélio~Renard Lavaud, Marie-Anne Lachaux, Pierre Stock, Teven~Le Scao, Thibaut Lavril, Thomas Wang, Timothée Lacroix, and William~El Sayed.
\newblock Mistral 7b, 2023.

\bibitem[Kil et~al.(2024)Kil, Tavazoee, Kang, and Kim]{kil-etal-2024-ii}
Jihyung Kil, Farideh Tavazoee, Dongyeop Kang, and Joo-Kyung Kim.
\newblock {II}-{MMR}: Identifying and improving multi-modal multi-hop reasoning in visual question answering.
\newblock In Lun-Wei Ku, Andre Martins, and Vivek Srikumar (eds.), \emph{Findings of the Association for Computational Linguistics: ACL 2024}, pp.\  10698--10709, Bangkok, Thailand, August 2024. Association for Computational Linguistics.
\newblock \doi{10.18653/v1/2024.findings-acl.636}.
\newblock URL \url{https://aclanthology.org/2024.findings-acl.636}.

\bibitem[Lerner et~al.(2022)Lerner, Ferret, Guinaudeau, Le~Borgne, Besan{\c{c}}on, Moreno, and Lov{\'o}n~Melgarejo]{lerner2022viquae}
Paul Lerner, Olivier Ferret, Camille Guinaudeau, Herv{\'e} Le~Borgne, Romaric Besan{\c{c}}on, Jos{\'e}~G Moreno, and Jes{\'u}s Lov{\'o}n~Melgarejo.
\newblock Viquae, a dataset for knowledge-based visual question answering about named entities.
\newblock In \emph{Proceedings of the 45th International ACM SIGIR Conference on Research and Development in Information Retrieval}, pp.\  3108--3120, 2022.

\bibitem[Lin \& Byrne(2022)Lin and Byrne]{lin2022retrieval}
Weizhe Lin and Bill Byrne.
\newblock Retrieval augmented visual question answering with outside knowledge.
\newblock In \emph{Proceedings of the 2022 Conference on Empirical Methods in Natural Language Processing}, pp.\  11238--11254, 2022.

\bibitem[Lin et~al.(2024)Lin, Chen, Mei, Coca, and Byrne]{lin2024fine}
Weizhe Lin, Jinghong Chen, Jingbiao Mei, Alexandru Coca, and Bill Byrne.
\newblock Fine-grained late-interaction multi-modal retrieval for retrieval augmented visual question answering.
\newblock \emph{Advances in Neural Information Processing Systems}, 36, 2024.

\bibitem[Lin et~al.(2022)Lin, Xie, Chen, Xu, Zhu, and Yuan]{lin2022revive}
Yuanze Lin, Yujia Xie, Dongdong Chen, Yichong Xu, Chenguang Zhu, and Lu~Yuan.
\newblock Revive: Regional visual representation matters in knowledge-based visual question answering.
\newblock \emph{Advances in Neural Information Processing Systems}, 35:\penalty0 10560--10571, 2022.

\bibitem[Liu et~al.(2024{\natexlab{a}})Liu, Xue, Chen, Chen, Zhao, Wang, Hou, Li, and Peng]{liu2024survey}
Hanchao Liu, Wenyuan Xue, Yifei Chen, Dapeng Chen, Xiutian Zhao, Ke~Wang, Liping Hou, Rongjun Li, and Wei Peng.
\newblock A survey on hallucination in large vision-language models.
\newblock \emph{arXiv preprint arXiv:2402.00253}, 2024{\natexlab{a}}.

\bibitem[Liu et~al.(2024{\natexlab{b}})Liu, Li, Li, and Lee]{liu2024improved}
Haotian Liu, Chunyuan Li, Yuheng Li, and Yong~Jae Lee.
\newblock Improved baselines with visual instruction tuning, 2024{\natexlab{b}}.

\bibitem[Liu et~al.(2024{\natexlab{c}})Liu, Li, Wu, and Lee]{liu2024visual}
Haotian Liu, Chunyuan Li, Qingyang Wu, and Yong~Jae Lee.
\newblock Visual instruction tuning.
\newblock \emph{Advances in neural information processing systems}, 36, 2024{\natexlab{c}}.

\bibitem[Loshchilov \& Hutter(2019)Loshchilov and Hutter]{loshchilov2018decoupled}
Ilya Loshchilov and Frank Hutter.
\newblock Decoupled weight decay regularization.
\newblock In \emph{International Conference on Learning Representations}, 2019.

\bibitem[Lu et~al.(2024{\natexlab{a}})Lu, Liu, Zhang, Wang, Dong, Liu, Sun, Ren, Li, Sun, et~al.]{lu2024deepseek}
Haoyu Lu, Wen Liu, Bo~Zhang, Bingxuan Wang, Kai Dong, Bo~Liu, Jingxiang Sun, Tongzheng Ren, Zhuoshu Li, Yaofeng Sun, et~al.
\newblock Deepseek-vl: towards real-world vision-language understanding.
\newblock \emph{arXiv preprint arXiv:2403.05525}, 2024{\natexlab{a}}.

\bibitem[Lu et~al.(2024{\natexlab{b}})Lu, Liu, Zhang, Wang, Dong, Liu, Sun, Ren, Li, Yang, Sun, Deng, Xu, Xie, and Ruan]{lu2024deepseekvl}
Haoyu Lu, Wen Liu, Bo~Zhang, Bingxuan Wang, Kai Dong, Bo~Liu, Jingxiang Sun, Tongzheng Ren, Zhuoshu Li, Hao Yang, Yaofeng Sun, Chengqi Deng, Hanwei Xu, Zhenda Xie, and Chong Ruan.
\newblock Deepseek-vl: Towards real-world vision-language understanding, 2024{\natexlab{b}}.

\bibitem[Marino et~al.(2019)Marino, Rastegari, Farhadi, and Mottaghi]{marino2019ok}
Kenneth Marino, Mohammad Rastegari, Ali Farhadi, and Roozbeh Mottaghi.
\newblock Ok-vqa: A visual question answering benchmark requiring external knowledge.
\newblock In \emph{Proceedings of the IEEE/cvf conference on computer vision and pattern recognition}, pp.\  3195--3204, 2019.

\bibitem[Radford et~al.(2021)Radford, Kim, Hallacy, Ramesh, Goh, Agarwal, Sastry, Askell, Mishkin, Clark, et~al.]{radford2021learning}
Alec Radford, Jong~Wook Kim, Chris Hallacy, Aditya Ramesh, Gabriel Goh, Sandhini Agarwal, Girish Sastry, Amanda Askell, Pamela Mishkin, Jack Clark, et~al.
\newblock Learning transferable visual models from natural language supervision.
\newblock In \emph{International conference on machine learning}, pp.\  8748--8763. PMLR, 2021.

\bibitem[Schwenk et~al.(2022)Schwenk, Khandelwal, Clark, Marino, and Mottaghi]{schwenk2022okvqa}
Dustin Schwenk, Apoorv Khandelwal, Christopher Clark, Kenneth Marino, and Roozbeh Mottaghi.
\newblock A-okvqa: A benchmark for visual question answering using world knowledge.
\newblock In \emph{European conference on computer vision}, pp.\  146--162. Springer, 2022.

\bibitem[Sun et~al.(2023)Sun, Shen, Cao, Liu, Li, Shen, Gan, Gui, Wang, Yang, et~al.]{sun2023aligning}
Zhiqing Sun, Sheng Shen, Shengcao Cao, Haotian Liu, Chunyuan Li, Yikang Shen, Chuang Gan, Liang-Yan Gui, Yu-Xiong Wang, Yiming Yang, et~al.
\newblock Aligning large multimodal models with factually augmented rlhf.
\newblock \emph{arXiv preprint arXiv:2309.14525}, 2023.

\bibitem[Vu et~al.(2023)Vu, Iyyer, Wang, Constant, Wei, Wei, Tar, Sung, Zhou, Le, et~al.]{vu2023freshllms}
Tu~Vu, Mohit Iyyer, Xuezhi Wang, Noah Constant, Jerry Wei, Jason Wei, Chris Tar, Yun-Hsuan Sung, Denny Zhou, Quoc Le, et~al.
\newblock Freshllms: Refreshing large language models with search engine augmentation.
\newblock \emph{arXiv preprint arXiv:2310.03214}, 2023.

\bibitem[Wang et~al.(2023)Wang, Lv, Yu, Hong, Qi, Wang, Ji, Yang, Zhao, Song, et~al.]{wang2023cogvlm}
Weihan Wang, Qingsong Lv, Wenmeng Yu, Wenyi Hong, Ji~Qi, Yan Wang, Junhui Ji, Zhuoyi Yang, Lei Zhao, Xixuan Song, et~al.
\newblock Cogvlm: Visual expert for pretrained language models.
\newblock \emph{arXiv preprint arXiv:2311.03079}, 2023.

\bibitem[Wei et~al.(2022)Wei, Wang, Schuurmans, Bosma, Xia, Chi, Le, Zhou, et~al.]{wei2022chain}
Jason Wei, Xuezhi Wang, Dale Schuurmans, Maarten Bosma, Fei Xia, Ed~Chi, Quoc~V Le, Denny Zhou, et~al.
\newblock Chain-of-thought prompting elicits reasoning in large language models.
\newblock \emph{Advances in neural information processing systems}, 35:\penalty0 24824--24837, 2022.

\bibitem[Wu et~al.(2023)Wu, Gan, Chen, Wan, and Philip]{wu2023multimodal}
Jiayang Wu, Wensheng Gan, Zefeng Chen, Shicheng Wan, and S~Yu Philip.
\newblock Multimodal large language models: A survey.
\newblock In \emph{2023 IEEE International Conference on Big Data (BigData)}, pp.\  2247--2256. IEEE, 2023.

\bibitem[Wu et~al.(2024)Wu, Yang, Chai, Zhang, Liu, Du, Liang, Shu, Cheng, Sun, et~al.]{wu2024tablebench}
Xianjie Wu, Jian Yang, Linzheng Chai, Ge~Zhang, Jiaheng Liu, Xinrun Du, Di~Liang, Daixin Shu, Xianfu Cheng, Tianzhen Sun, et~al.
\newblock Tablebench: A comprehensive and complex benchmark for table question answering.
\newblock \emph{arXiv preprint arXiv:2408.09174}, 2024.

\bibitem[Xu et~al.(2023)Xu, Peng, Lei, Mukherjee, Liu, and Xu]{xu2023rewoo}
Binfeng Xu, Zhiyuan Peng, Bowen Lei, Subhabrata Mukherjee, Yuchen Liu, and Dongkuan Xu.
\newblock Rewoo: Decoupling reasoning from observations for efficient augmented language models.
\newblock \emph{arXiv preprint arXiv:2305.18323}, 2023.

\bibitem[Yang et~al.(2022)Yang, Gan, Wang, Hu, Lu, Liu, and Wang]{yang2022empirical}
Zhengyuan Yang, Zhe Gan, Jianfeng Wang, Xiaowei Hu, Yumao Lu, Zicheng Liu, and Lijuan Wang.
\newblock An empirical study of gpt-3 for few-shot knowledge-based vqa.
\newblock In \emph{Proceedings of the AAAI conference on artificial intelligence}, volume~36, pp.\  3081--3089, 2022.

\bibitem[Yao et~al.(2023)Yao, Zhao, Yu, Du, Shafran, Narasimhan, and Cao]{yao2023react}
Shunyu Yao, Jeffrey Zhao, Dian Yu, Nan Du, Izhak Shafran, Karthik Narasimhan, and Yuan Cao.
\newblock React: Synergizing reasoning and acting in language models.
\newblock In \emph{International Conference on Learning Representations (ICLR)}, 2023.

\bibitem[Ye et~al.(2024)Ye, Xu, Xu, Ye, Yan, Zhou, Wang, Hu, Shi, Shi, Li, Xu, Chen, Tian, Qian, Zhang, Huang, and Zhou]{ye2024mplugowl}
Qinghao Ye, Haiyang Xu, Guohai Xu, Jiabo Ye, Ming Yan, Yiyang Zhou, Junyang Wang, Anwen Hu, Pengcheng Shi, Yaya Shi, Chenliang Li, Yuanhong Xu, Hehong Chen, Junfeng Tian, Qi~Qian, Ji~Zhang, Fei Huang, and Jingren Zhou.
\newblock mplug-owl: Modularization empowers large language models with multimodality, 2024.

\bibitem[Yin et~al.(2023)Yin, Fu, Zhao, Li, Sun, Xu, and Chen]{yin2023survey}
Shukang Yin, Chaoyou Fu, Sirui Zhao, Ke~Li, Xing Sun, Tong Xu, and Enhong Chen.
\newblock A survey on multimodal large language models.
\newblock \emph{arXiv preprint arXiv:2306.13549}, 2023.

\bibitem[Yu et~al.(2024)Yu, Yao, Zhang, He, Han, Cui, Hu, Liu, Zheng, Sun, et~al.]{yu2024rlhf}
Tianyu Yu, Yuan Yao, Haoye Zhang, Taiwen He, Yifeng Han, Ganqu Cui, Jinyi Hu, Zhiyuan Liu, Hai-Tao Zheng, Maosong Sun, et~al.
\newblock Rlhf-v: Towards trustworthy mllms via behavior alignment from fine-grained correctional human feedback.
\newblock In \emph{Proceedings of the IEEE/CVF Conference on Computer Vision and Pattern Recognition}, pp.\  13807--13816, 2024.

\bibitem[Zhai et~al.(2023)Zhai, Yang, Zhao, Xu, Shen, Zhao, Keutzer, Li, Yan, and Fan]{zhai2023halle}
Bohan Zhai, Shijia Yang, Xiangchen Zhao, Chenfeng Xu, Sheng Shen, Dongdi Zhao, Kurt Keutzer, Manling Li, Tan Yan, and Xiangjun Fan.
\newblock Halle-switch: Rethinking and controlling object existence hallucinations in large vision language models for detailed caption.
\newblock \emph{arXiv preprint arXiv:2310.01779}, 2023.

\bibitem[Zhang et~al.(2023)Zhang, Yao, Zhang, Tang, Ma, He, Wang, Gerstein, Wang, Liu, et~al.]{zhang2023igniting}
Zhuosheng Zhang, Yao Yao, Aston Zhang, Xiangru Tang, Xinbei Ma, Zhiwei He, Yiming Wang, Mark Gerstein, Rui Wang, Gongshen Liu, et~al.
\newblock Igniting language intelligence: The hitchhiker's guide from chain-of-thought reasoning to language agents.
\newblock \emph{arXiv preprint arXiv:2311.11797}, 2023.

\bibitem[Zhao et~al.(2024)Zhao, Zhang, Yu, Wang, Geng, Fu, Yang, Zhang, and Cui]{zhao2024retrieval}
Penghao Zhao, Hailin Zhang, Qinhan Yu, Zhengren Wang, Yunteng Geng, Fangcheng Fu, Ling Yang, Wentao Zhang, and Bin Cui.
\newblock Retrieval-augmented generation for ai-generated content: A survey.
\newblock \emph{arXiv preprint arXiv:2402.19473}, 2024.

\bibitem[Zhao et~al.(2023)Zhao, Chen, Wang, Jiao, Qin, Ding, Guo, Li, Li, Joty, et~al.]{zhao2023retrieving}
Ruochen Zhao, Hailin Chen, Weishi Wang, Fangkai Jiao, Chengwei Qin, Bosheng Ding, Xiaobao Guo, Minzhi Li, Xingxuan Li, Shafiq Joty, et~al.
\newblock Retrieving multimodal information for augmented generation: A survey.
\newblock In \emph{The 2023 Conference on Empirical Methods in Natural Language Processing}, 2023.

\end{thebibliography}
\bibliographystyle{iclr2025_conference}

\newpage
\appendix
\section{More Details on Dyn-VQA Dataset}
\subsection{Statistics}

\begin{figure}[th]
\centering
\scalebox{0.9}{
\includegraphics[width=0.45\textwidth]{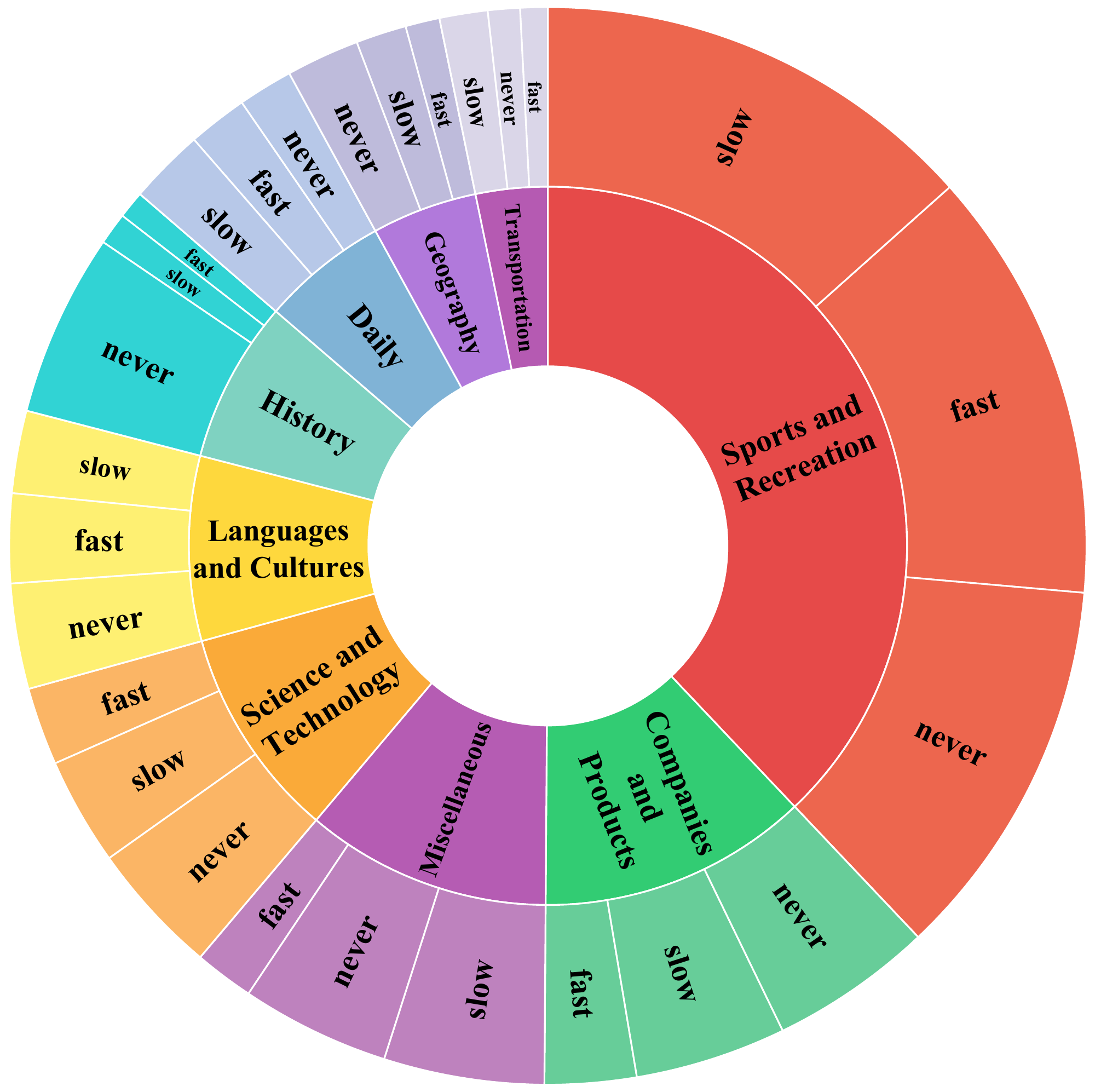}}
\caption{Data distribution of different domains and answer change frequencies on the Dyn-VQA dataset.}
\label{fig:data_dist}
\end{figure}

Figure \ref{fig:data_dist} illustrates the data distribution across various domains and the answer change frequencies in Dyn-VQA. Among the 9 domains, Sports and Recreation, and Companies and Products constitute approximately 50\% of the data. The distribution of questions with fast, slow, and never answer changes is relatively balanced within the classes and does not exhibit a long tail, reflecting a distribution that closely aligns with real-world scenarios.

\subsection{Dataset Quality}
To ensure dataset quality, following the initial annotation of Dyn-VQA, we employed two quality control (QC) annotators to re-evaluate the dataset. This re-evaluation included verification of the answers, domains, answer update frequencies, reasoning steps, and whether require external visual knowledge for each data instance. Data deemed incorrect by both QC annotators was filtered out. The agreement between QC annotators \#1 and \#2 with the initial annotations, as well as the agreement between the two QC annotators, are presented in Table \ref{tab:annotation}. The inter-annotator agreement measured by Fleiss’s Kappa \citep{fleiss1971measuring} all exceeded 0.8, demonstrating the reliability of the annotation results. 

\begin{table*}[h!]
\centering
\begin{minipage}[t]{0.6\textwidth} 
\centering
\caption{The inter-annotator agreement of QC\#1, QC\#2 and initial annotation with each other.}
\scalebox{0.67}{
\begin{tabular}{lccccc}
\toprule
                      & \textbf{Answer} & \textbf{Domain} & \begin{tabular}[c]{@{}c@{}}\textbf{Answer}\\\textbf{Update Freq.}\end{tabular}  & \begin{tabular}[c]{@{}c@{}}\textbf{Reasoning}\\\textbf{Step}\end{tabular}     & \begin{tabular}[c]{@{}c@{}}\textbf{External}\\\textbf{Visual-Seek}\end{tabular} \\ \cmidrule{1-6}
Init. vs. QC\#1       & 81.2     & 84.4     & 89.6                &  84.1            & 87.9                      \\
Init. vs. QC\#2       & 84.2     & 85.6     & 85.4               & 82.3           & 84.5                      \\
QC\#1 vs. QC\#2      & 80.1     & 83.8                & 86.9             & 81.7 & 85.0                     \\ \cmidrule{1-6}
Avg.                   & 81.8       & 84.6       & 87.3                  & 82.7               & 85.8          \\ \bottomrule              
\end{tabular}}
\label{tab:annotation}
\end{minipage}%
\hspace{0.1cm}
\begin{minipage}[t]{0.38\textwidth} 
\centering
\caption{Question and answer diversity. Mean pairwise cosine distances are used as metrics.}
\scalebox{0.8}{
\begin{tabular}{lcc}
\toprule
\textbf{Dataset} & \textbf{Question} & \textbf{Answer} \\ \midrule
VQAv2            & 0.8405            & 0.7606          \\
A-OKVQA          & 0.8428            & 0.9078          \\
InfoSeek         & 0.7569            & 0.8918          \\ \midrule
Dyn-VQA          & 0.8532            & 0.9135          \\ \bottomrule
\end{tabular}}
\label{tab:data_diverse}
\end{minipage}

\end{table*}

\subsection{Dataset Diversity}
To assess the diversity of Dyn-VQA compared with other datasets, we calculated pairwise cosine distances for each dataset. Following A-OKVQA, we utilized a sentence-transformers\footnote{\url{https://huggingface.co/sentence-transformers/multi-qa-MiniLM-L6-cos-v1}} as the encoder. As indicated in Table \ref{tab:data_diverse}, Dyn-VQA exhibits more diverse data types, evidenced by larger cosine distances. Intuitively, questions in InfoSeek are primarily constructed through templates, resulting in more homogeneous data format. In contrast, Dyn-VQA is manually curated. Moreover, since the questions of Dyn-VQA are all from open-ended domains, the answers of them feature longer response length compared with the same manually constructed VQAv2 and A-OKVQA, which can essentially be answered with a single word.

\subsection{Regular Updates to Dyn-VQA}
The biggest characteristic of Dyn-VQA is that the knowledge required to answer the questions it contains is dynamically updated over time, that is, the answers to the questions in Dyn-VQA are constantly changing. Therefore, to ensure that Dyn-VQA can serve as an effective research resource for the community in the long term, it is necessary for us to dynamically update and maintain Dyn-VQA's answer annotation information. \textbf{In general, based on the analysis of the frequency of change of Dyn-VQA's answer, we commit to updating Dyn-VQA's answer annotation every three months to ensure its timeliness.}  

In particular, we plan to implement a semi-automatic data update mechanism in which models and humans work together. Specifically, For a certain sample in Dyn-VQA, we first retrieve relevant text knowledge through the search engine, and then we use LLMs such as Qwen1.5-72B to compare the latest knowledge retrieved with the original answer to determine whether the original answer for the sample needs to be updated. Note that here we only need LLMs to determine whether the answer needs to be updated, and do not need them to accurately answer the latest answer, because we think this is simpler and more friendly for LLMs. After we have the prediction results of LLMs as a basis, we then carry out the manual update process. We require every human annotator involved in the data update process to accurately update each sample's answer based on their common sense, the results of large model judgments, and the latest relevant knowledge retrieved by the search engines.
We believe that the semi-automatic update mechanism, in which models and humans cooperate, will not only reduce the workload of manual annotators but also improve the accuracy of data updates.

\section{Related Works}
\subsection{Multimodal Large Language Models}
In 2023, with the advent of GPT-4V \citep{achiam2023gpt}, a series of MLLMs \citep{Qwen-VL,lu2024deepseek, instructblip, liu2024visual, wang2023cogvlm} have been proposed and demonstrated superior results on a variety of vision-language tasks \citep{yin2023survey, wu2023multimodal}. Despite promising results, MLLMs tend to haphazardly produce responses that appear plausible yet contain factual errors when faced with real-world questions. Many prior works explore mitigating this hallucination issue \citep{liu2024survey,  bai2024hallucination} by introducing additional knowledge-enhancing data or tasks into the different training stages of MLLMs, including pre-training \citep{zhai2023halle}, instruction fine-tuning \citep{xcot, jain2024vcoder}, and RLHF \citep{sun2023aligning, yu2024rlhf}. However, the expensive training cost of MLLMs pose significant challenges to the scalability of these methods. Therefore, mRAG attracts growing interest as an effective and efficient alternative.

\subsection{Multimodal Retrieval Augmented Generation}
Besides retrieval from knowledge bases as described in Related Works section in main content, some work explores retrieval from other knowledge sources. REVEAL \citep{hu2023reveal} integrates knowledge retrieved from multiple sources, including wikidata, wikipedia and other VQA datasets. PICa \citep{yang2022empirical} consider LLMs as implicit knowledge bases and extract relevant information from GPT-3 \citep{brown2020language} with image description as prompt.

\subsection{VQA datasets as mRAG Benchmarks}
Knowledge-seeking VQA datasets \citep{marino2019ok,jain2021select,schwenk2022okvqa,kil-etal-2024-ii, chen2024measuring} are widely employed to evaluate the performance of mRAGs, which rely on external information to address open-domain visual questions. For instance, the recently introduced Wikipedia-based VQA dataset, InfoSeek \citep{chen2023can}, emphasizes fine-grained entity knowledge for open-domain questions. A-OKVQA \citep{schwenk2022okvqa} is a new knowledge-based VQA benchmark that necessitates a broad spectrum of commonsense and world knowledge. As illustrated in Table \ref{tab:data_compare}, several knowledge-seeking VQA datasets have been proposed in recent years.

Nevertheless, the knowledge scope assessed by these datasets is constrained. More critically, the questions in these datasets often exhibit a fixed format, typically querying a specific property of the object in image, which is generally textual knowledge available on the internet. Such two-hop questions can be readily addressed by standard two-step retrieval. The static nature of these datasets prompts us to propose Dyn-VQA, which requires the retrieval of dynamic knowledge, knowledge from a more diverse range of modalities, or more complex multi-hop knowledge.

\section{Experiments}
\subsection{Baselines}
In the main experiment of Table \ref{tab:main}, we also introduced generative search engine and human as baseline models. We describe them following:

\noindent\textbf{Generative Search Engine} 
Among the commercial AI products, LLM-powered generative search engines like Bing Chat, PerplexityAI, mita.ai, Tongyi, and GPT-4o stand out. For our experiment, we select Bing Chat, pro-version of PerplexityAI, and Gemini-Advance as representatives of AI search engines. They have multimodal RAG ability for fair comparison.

\noindent\textbf{Human Performance.} We also investigated human performance on Dyn-VQA, employing participants with at least a bachelor degree. These participants were not involved in the Dyn-VQA annotation process.
\subsection{Other Backbone MLLMs for Heuristic mRAGs}
In the Appendix, we supplement the main experiment with more MLLM as backbone for heuristic mRAGs. \textbf{Deepseek-VL-7B-Chat} is an open-source large visual language model introduced by \citet{lu2024deepseekvl}. \textbf{VisualGLM-6B} is an open-source, bilingual, multi-modal large visual language model proposed by \citet{du2022glm}. \textbf{Llava-V1.6-Mistral-7B} is an open-source large visual language model proposed by \citet{liu2024improved}, which is trained by fine-tuning LLM on multimodal instruction-following data based on \citet{jiang2023mistral}. \textbf{mPLUG-Owl2.1} \citep{ye2024mplugowl} is a large visual language model trained with a two-stage method for aligning image and text.

\subsection{Training Details}
OmniSearch (G) is constructed by prompt engineering for GPT-4V, whose prompt template can be found in the next subsection. OmniSearch (Q) is developed by instruction fine-tuning of Qwen-VL-Chat. we synthesize instruction data containing planning trajectories by using GPT-4V and raw InfoSeek data. $\sim$40K data is synthesized, and then the data is filtered by a sequence of predefined rules. 13K data is eventually obtained. Additionally, general instruction data from CogVLM-SFT-311K\footnote{\url{https://huggingface.co/datasets/THUDM/CogVLM-SFT-311K}} is mixed in a ratio of 1:2 with the planning instruction data. We use LoRA to parameter-efficient fine-tune Qwen-VL-Chat, with LoRA rank and alpha are 8 and 32. AdamW \citep{loshchilov2018decoupled} optimizer is employed for model training, with learning rate set of 1e-4 and weight decay of 0.1. We utilize a cosine learning rate schedule, warming up over 5\% of the training steps. The model is fine-tuned with 1 epochs, with the batch size per device set to 4 and the gradient accumulation step set to 8. The maximum sequence length is 8192. The training are run on 4 NVIDIA A100 SXM4 80GB GPUs.

All data is in the form of multi-round conversations. During model training, we expect the model to learn to generate response given the instruction and input text, thus we compute the loss function by considering only the response tokens of each round and ignoring the input tokens.

\subsection{Supplement to Main Experiments}
Experiment results of using more MLLMs as backbones for heuristic mRAGs are supplemented in Table \ref{tab:main_appendix}, from which we can find:

(1) Among all open-sourced MLLMs, Deepseek-VL-7B-Chat exhibits the best performance as a backbone, achieving the highest overall performance. While heuristic mRAG methods show the most significant improvements with InstructBLIP-Vicuna-7B, where the average absolute gain across four heuristic mRAG approaches reaches 16.60, marking an enhancement of 223.1\% compared to the baseline performance of InstructBLIP-Vicuna-7B. This substantial increase is likely attributed to the initially lower performance of the original InstructBLIP-Vicuna-7B.

(2) The performance variance between MLLMs that enhanced by the heuristic mRAG becomes smaller. For instance, with the incorporation of the image caption-based two-step mRAG method, the variance in performance among open-source MLLMs reduces from 4.92 to 2.42. Combined with the analysis in (1), this suggests that for MLLMs with suboptimal foundational capabilities, mRAG serves as an optimal method to bolster model performance. It is not only less resource-dependent, but also convenient to deploy.

(3) Although single-step heuristic mRAGs may not retrieve the most precise content, they still benefit the original model capabilities. Retrieving image information with input images enriches the MLLM with information about the objects in images, whereas web page retrieval with input questions can also retrieve some relevant information due to certain keywords in the questions. Overall, web page retrieval yields a greater improvement with an average of 6.78, which might be attributable to the inherent capacity of MLLMs to recognize some objects depicted in the images. It also indicates that unlike previous VQA datasets, the challenge of Dyn-VQA does not lie solely in the recognition of objects in images.

\section{Analysis Experiments on OmniSearch}

\begin{table*}[]
\centering
\caption{Experiments on the impact of using different parts of retrieved content. We report the performance of OmniSearch (G).}
\scalebox{0.65}{
\begin{tabular}{llcccccccccc}
\toprule
\multirow{2}{*}{\textbf{Retrieved Content}} & \multirow{2}{*}{\textbf{Using Part}} & \multicolumn{3}{c}{\textbf{Answer Update Frequency}} & \multicolumn{2}{c}{\textbf{Reasoning Steps}} & \multicolumn{2}{c}{\textbf{Visual-Seeking}} & \multicolumn{2}{c}{\textbf{Language}} & \multirow{2}{*}{\textbf{all}} \\ \cmidrule{3-11}
                                &                             & fast          & slow          & never          & $\leq$ 2-hop      & $>$ 2-hop          & no              & yes              & zh            & en           &                      \\ \midrule
\multirow{3}{*}{Image}          & Image \& Caption            & 44.04 & 49.58 & 54.45 & 50.38 & 49.06 & 50.49 & 49.73 & 46.96 & 53.21 & 50.03                \\
                                & - Caption                  & 41.65 & 47.23 & 52.03 & 48.00 & 46.67 & 48.02 & 47.39 & 44.57 & 50.81 & 47.64     \\
                                & - Image                & 42.58 & 48.14 & 52.97 & 48.93 & 47.60 & 48.98 & 48.30 & 45.50 & 51.74 & 48.57       \\ \midrule
\multirow{3}{*}{Web Snippet}    & Web Title \& Description    & 44.04 & 49.58 & 54.45 & 50.38 & 49.06 & 50.49 & 49.73 & 46.96 & 53.21  & 50.03       \\
                                & + Related Knowledge         &  45.26 & 50.78 & 55.68 & 51.60 & 50.28 & 51.75 & 50.92 & 48.17 & 54.43 & 51.25 \\
                                & - Web Title        &   42.09 & 47.66 & 52.47 & 48.44 & 47.11 & 48.47 & 47.82 & 45.01 & 51.25 & 48.08        \\ \bottomrule
\end{tabular}}
\label{tab:exp_ret_content}
\end{table*}

\subsection{Analysis Experiments on Retrieved Content}
\textbf{Is each part of the retrieved content useful?} 
Table \ref{tab:exp_ret_content} presents the impact of utilizing different parts of the content returned by the retrieval APIs, from which we can find that each part of the retrieved content is beneficial to overall performance. Utilizing different single part results in varying degrees of performance degradation compared to using the entire retrieved content. Notably, image captions contribute most to final performance. This phenomenon arises primarily because nearly all questions in Dyn-VQA necessitate object recognition in images, for which captions from retrieved similar images provide crucial additional information to the model. In contrast, the benefit derived from incorporating relevant knowledge provided by the search engine is relatively modest. This is predominantly due to such information typically constituting static background knowledge that lacks direct relevance to the specific problem at hand.

\begin{table*}[]
\centering
\caption{Main results on Dyn-VQA.}
\scalebox{0.72}{
\begin{tabular}{lcccccccccc}
\toprule
\multirow{2}{*}{\textbf{Model}}   & \multicolumn{3}{c}{\textbf{Answer Update Frequency}}                     & \multicolumn{2}{c}{\textbf{Reasoning Steps}}         & \multicolumn{2}{c}{\textbf{Visual-Seeking}}          & \multicolumn{2}{c}{\textbf{Language}}                & \multirow{2}{*}{\textbf{all}} \\ \cmidrule{2-10}
                        & fast          & slow          & never          & $\leq$ 2-hop      & $>$ 2-hop     & no              & yes              & zh            & en           &                      \\ \midrule
\multicolumn{11}{l}{\textbf{\textit{Original (M)LLMs} }}                                   \\
Qwen-VL-Chat             &  13.69 & 14.20 & 17.27 & 15.56 & 14.49 & 15.50 & 15.13 & 16.92 & 13.58 & 15.28                  \\
Qwen-7B-Chat             & 5.63 & 7.86 & 15.97 & 10.48 & 10.43 & 11.05 & 10.08 & 10.48 & 10.46 & 10.47              \\
Deepseek-VL-7B-Chat    & 11.66 & 18.57 & 30.16 & 22.12 & 19.08 & 23.26 & 19.99 & 19.94 & 22.73 & 21.31           \\
VisualGLM-6B             &  12.05 & 13.94 & 20.92 & 16.63 & 14.99 & 18.92 & 14.34 & 16.21 & 16.18 & 16.19               \\
Llava-V1.6-Mistral-7B    &   15.39 & 19.72 & 28.34 & 22.65 & 20.12 & 23.53 & 20.92 & 16.27 & 27.85 & 21.97                 \\
mPLUG-Owl2.1             &   8.82  & 11.25 & 18.44 & 13.81 & 12.44 & 14.48 & 12.74 & 8.05  & 19.01 & 13.44                  \\
InstructBLIP-Vicuna-7B   &   6.08  & 7.40  & 8.38  & 7.48  & 7.33  & 7.51  & 7.38  & 4.65  & 10.31 & 7.44                   \\
Qwen-VL-Max              &    15.11 & 30.44 & 39.51 & 30.22 & 29.21 & 31.10 & 29.18 & 22.81 & 37.32 & 29.96               \\
GPT-4V                   &   17.63 & 27.80 & 40.82 & 30.80 & 28.74 & 31.71 & 29.26 & 26.44 & 34.18 & 30.25                  \\ \midrule
\multicolumn{11}{l}{\textbf{+ \textit{Heuristic mRAG: Retrieving Images with Input Images} }}                                   \\
Qwen-VL-Chat            & 15.74 & 17.12 & 25.74 & 20.52 & 19.14 & 22.68 & 18.44 & 22.06 & 18.19 & 20.16                \\
Qwen-7B-Chat            & 10.97 & 15.04 & 25.91 & 18.76 & 16.86 & 24.18 & 14.23 & 16.8 & 19.75 & 18.25              \\
Deepseek-VL-7B-Chat      & 15.95 & 22.63 & 36.80 & 26.49 & 26.35 & 32.27 & 22.50 & 25.97 & 26.94 & 26.45               \\
VisualGLM-6B            & 15.51 & 16.69 & 29.08 & 21.73 & 19.99 & 25.75 & 18.22 & 20.84 & 21.71 & 21.27              \\
Llava-V1.6-Mistral-7B    & 17.38 & 19.74 & 33.48 & 25.42 & 22.10 & 32.00 & 19.47 & 17.01 & 32.29 & 24.54              \\
mPLUG-Owl2.1            & 13.49 & 16.52 & 28.14 & 21.42 & 17.20 & 25.84 & 16.54 & 13.45 & 27.36 & 20.30            \\
InstructBLIP-Vicuna-7B  & 13.40 & 15.83 & 29.02 & 20.19 & 19.73 & 25.15 & 16.68 & 13.10 & 25.05 & 20.07         \\
Qwen-VL-Max              &  24.04 & 28.99 & 45.49 & 34.22 & 34.08 & 41.54 & 29.19 & 31.07 & 37.39 & 34.19                 \\
GPT-4V                  & 20.18 & 33.21 & 50.00 & 35.94 & 35.65 & 42.63 & 31.33 & 30.90 & 40.32 & 35.87              \\ \midrule
\multicolumn{11}{l}{\textbf{+ \textit{Heuristic mRAG: Retrieving Web Pages with Input Questions} }}                                   \\
Qwen-VL-Chat            & 20.78 & 18.27 & 27.61 & 22.76 & 22.20 & 23.34 & 22.07 & 26.66 & 17.94 & 22.59               \\
Qwen-7B-Chat             & 14.65 & 15.47 & 24.98 & 19.14 & 18.66 & 19.99 & 18.34 & 17.93 & 20.12 & 19.01         \\
Deepseek-VL-7B-Chat     & 20.34 & 23.44 & 32.02 & 26.13 & 25.65 & 27.48 & 25.01 & 25.03 & 27.01 & 26.00           \\
VisualGLM-6B            & 20.71 & 18.02 & 28.56 & 23.13 & 22.23 & 23.47 & 22.50 & 22.53 & 23.26 & 22.89              \\
Llava-V1.6-Mistral-7B    & 20.39 & 21.62 & 30.33 & 24.99 & 24.01 & 26.70 & 23.39 & 20.96 & 28.62 & 24.73           \\
mPLUG-Owl2.1           & 20.49 & 24.29 & 30.93 & 26.04 & 25.53 & 28.42 & 24.19 & 21.16 & 30.79 & 25.90                 \\
InstructBLIP-Vicuna-7B   & 21.63 & 18.79 & 27.65 & 23.44 & 21.95 & 23.07 & 23.02 & 19.98 & 26.19 & 23.04        \\
Qwen-VL-Max              & 26.71 & 27.37 & 35.84 & 30.65 & 30.22 & 31.14 & 30.13 & 30.27 & 30.82 & 30.54           \\
GPT-4V                   & 22.48 & 30.92 & 40.84 & 33.00 & 31.47 & 34.32 & 31.42 & 31.10 & 34.13 & 32.59                    \\ \midrule
\multicolumn{11}{l}{\textbf{+ \textit{Heuristic Two-Step mRAG: Retrieving Image First, then Retrieving Web Pages with Question appended to Retrieved Caption} }}                                   \\
Qwen-VL-Chat            & 19.17 & 20.02 & 28.54 & 23.33 & 22.68 & 23.84 & 22.69 & 24.11 & 22.17 & 23.16        \\
Qwen-7B-Chat            & 15.27 & 17.33 & 28.53 & 21.70 & 19.83 & 26.65 & 17.50 & 20.47 & 21.96 & 21.20         \\
Deepseek-VL-7B-Chat     & 18.65 & 23.51 & 34.68 & 26.66 & 26.54 & 29.74 & 24.51 & 25.70 & 27.59 & 26.63         \\
VisualGLM-6B             & 18.56 & 19.98 & 29.43 & 23.87 & 21.84 & 27.07 & 20.79 & 22.00 & 24.70 & 23.33            \\
Llava-V1.6-Mistral-7B   & 18.41 & 20.81 & 34.70 & 26.27 & 23.97 & 30.49 & 22.37 & 19.10 & 32.41 & 25.65              \\
mPLUG-Owl2.1           & 14.10 & 19.28 & 30.24 & 22.77 & 20.74 & 26.96 & 19.02 & 16.24 & 28.41 & 22.23       \\
InstructBLIP-Vicuna-7B  & 16.22 & 18.95 & 30.34 & 22.96 & 22.07 & 25.47 & 20.86 & 16.46 & 29.18 & 22.72   \\
Qwen-VL-Max              &   24.44 & 30.75 & 43.21 & 34.03 & 33.91 & 38.26 & 31.1 & 32.04 & 36.01 & 33.99                \\
GPT-4V                 & 20.37 & 33.98 & 48.46 & 36.12 & 36.04 & 40.19 & 33.32 & 32.99 & 39.30 & 36.10         \\ \midrule
\multicolumn{11}{l}{\textbf{+ \textit{Heuristic Two-Step mRAG: Image Caption First, then Retrieving Web Pages with Question appended to Caption} }}                                   \\
Qwen-VL-Chat            & 22.05 & 25.87 & 31.84 & 27.58 & 26.21 & 27.44 & 27.06 & 28.81 & 25.57 & 27.21             \\
Qwen-7B-Chat            & 14.65 & 21.16 & 28.66 & 22.89 & 21.02 & 23.64 & 21.55 & 16.57 & 28.39 & 22.39              \\
Deepseek-VL-7B-Chat      & 21.12 & 27.65 & 36.27 & 29.41 & 29.08 & 29.72 & 29.05 & 26.84 & 31.88 & 29.32            \\
VisualGLM-6B            & 19.60 & 21.75 & 33.23 & 25.88 & 25.23 & 27.27 & 24.65 & 22.95 & 28.56 & 25.71           \\
Llava-V1.6-Mistral-7B   & 21.09 & 26.41 & 33.87 & 28.20 & 27.25 & 29.76 & 26.71 & 21.66 & 34.42 & 27.94           \\
mPLUG-Owl2.1            & 20.46 & 26.67 & 34.91 & 28.77 & 26.90 & 28.31 & 28.25 & 20.33 & 36.47 & 28.27             \\
InstructBLIP-Vicuna-7B   & 24.37 & 28.22 & 35.62 & 30.51 & 29.85 & 30.77 & 30.06 & 22.98 & 35.57 & 30.33           \\
Qwen-VL-Max              &  24.27 & 32.93 & 44.03 & 35.04 & 34.97 & 35.16 & 34.92 & 31.1 & 39.05 & 35.02                    \\
GPT-4V                  & 24.90 & 36.74 & 45.76 & 37.23 & 36.94 & 37.82 & 36.70 & 31.65 & 42.81 & 37.15            \\ \midrule
\multicolumn{11}{l}{\textbf{\textit{Generative Search Engine} }}                                   \\
Bing Chat                & 27.71 & 32.77 & 32.99 & 31.67 & 30.80 & 35.44 & 28.64 & 29.62 & 32.74 & 31.44                   \\
Perplexity AI            &   29.62 & 34.69 & 34.88 & 33.67 & 32.81 & 37.46 & 30.67 & 31.59 & 34.80 & 33.51            \\ 
Gemini                  & 36.17 & 32.86 & 42.84 & 38.75 & 34.78 & 46.39 & 31.82 & 35.77 & 39.69 & 37.69  \\ \midrule
\multicolumn{11}{l}{\textbf{\textit{Ours} }}   \\
OmniSearch (Qwen-VL-Chat) &  35.16 & 40.89 & 45.52 & 41.34 & 40.81 & 42.56 & 40.28 & 39.22 & 43.23 & 41.20             \\
OmniSearch (GPT-4V)       & \textbf{44.04} & \textbf{49.58} & \textbf{54.45} & \textbf{50.38} & \textbf{49.06} & \textbf{50.49} & \textbf{49.73} & \textbf{46.96} & \textbf{53.21} & \textbf{50.03}                     \\ \midrule
\multicolumn{11}{l}{\textbf{\textit{Estimated Upper Bound: + Retrieving Web Pages with Gloden Query} }}   \\
Qwen-VL-Chat            & 37.46 & 46.52 & 52.18 & 46.73 & 45.28 & 47.94 & 45.27 & 43.88 & 48.90 & 46.35           \\
Qwen-7B-Chat            & 39.69 & 47.27 & 57.76 & 49.53 & 49.02 & 50.94 & 48.36 & 46.02 & 52.88 & 49.40     \\
Deepseek-VL-7B-Chat      & 35.89 & 46.00 & 54.29 & 46.84 & 45.92 & 49.20 & 44.81 & 45.41 & 47.80 & 46.59           \\
VisualGLM-6B             & 36.09 & 40.09 & 50.59 & 43.33 & 42.75 & 44.00 & 42.61 & 41.98 & 44.40 & 43.17           \\
Llava-V1.6-Mistral-7B    & 39.67 & 48.56 & 56.09 & 49.67 & 47.80 & 52.38 & 47.00 & 43.90 & 54.61 & 49.17            \\
mPLUG-Owl2.1             & 41.19 & 48.66 & 55.83 & 49.63 & 49.17 & 51.77 & 47.97 & 44.49 & 54.68 & 49.51             \\
InstructBLIP-Vicuna-7B   & 37.17 & 43.92 & 55.53 & 46.51 & 45.94 & 49.22 & 44.45 & 41.25 & 54.05 & 46.36            \\
Qwen-VL-Max              & 42.19 & 53.01 & 56.6 & 51.91 & 50.58 & 52.97 & 50.6 & 49.83 & 53.33 & 51.56                  \\
GPT-4V                  & 45.59 & 54.23 & 60.78 & 55.15 & 52.81 & 54.53 & 54.51 & 51.08 & 58.07 & 54.52             \\ \midrule
Human Performance                & 51.63 & 60.02 & 53.19 & 54.12 & 58.31 & 57.86 & 53.20 & 51.96 & 58.36 & 55.12 \\ \bottomrule              
\end{tabular}}
\label{tab:main_appendix}
\end{table*}

\begin{table*}[]
\centering
\caption{Performances of GPT-4V and OmniSearch(G) with different top-k retrieved content.}
\scalebox{0.75}{
\begin{tabular}{llcccccccccc}
\toprule
\multirow{2}{*}{\textbf{Model}} & \multirow{2}{*}{\textbf{\# R.C.}} & \multicolumn{3}{c}{\textbf{Answer Update Frequency}} & \multicolumn{2}{c}{\textbf{Reasoning Steps}} & \multicolumn{2}{c}{\textbf{Visual-Seeking}} & \multicolumn{2}{c}{\textbf{Language}} & \multirow{2}{*}{\textbf{all}} \\ \cmidrule{3-11}
                                &                                      & fast          & slow          & never          & $\leq$ 2-hop      & $>$ 2-hop          & no              & yes              & zh            & en           &                      \\ \midrule
GPT-4V        & None            &    17.63 & 27.80 & 40.82 & 30.80 & 28.74 & 31.71 & 29.26 & 26.44 & 34.18 & 30.25                \\
GPT-4V        & 1                               & 24.07 & 32.6  & 46.11 & 35.47 & 34.98 & 37.98 & 33.54 & 32.91 & 38.02 & 35.33  \\
GPT-4V        & 3                     &   26.66 & 37.7  & 47.65 & 39.05 & 38.02 & 40.01 & 37.94 & 37.37 & 40.21 & 38.78    \\
GPT-4V        & All                         &    24.90 & 36.74 & 45.76 & 37.23 & 36.94 & 37.82 & 36.70 & 31.65 & 42.81 & 37.15      \\ \midrule
OmniSearch (G)         & 1            &   35.16 & 40.89 & 45.52 & 41.34 & 40.81 & 42.56 & 40.28 & 39.22 & 43.23 & 41.20                    \\
OmniSearch (G)         & 3                     &   39.08 & 49.06 & 52.87 & 48.93 & 45.89 & 52.38 & 44.98 & 48.89 & 48.23 & 48.09     \\
OmniSearch (G)         & All                     &   44.04 & 49.58 & 54.45 & 50.38 & 49.06 & 50.49 & 49.73 & 46.96 & 53.21 & 50.03  \\ \bottomrule
\end{tabular}}
\label{tab:exp_topk}
\end{table*}

\begin{table*}[th]
\centering
\caption{Consistency of Different Metrics Consistency of Different Metrics Consistency of Different Metrics.}
\scalebox{1}{
\begin{tabular}{lcccccc}
\toprule
\multirow{2}{*}{\textbf{Model}}       & \multirow{2}{*}{\textbf{Recall}} & \multirow{2}{*}{\begin{tabular}[c]{@{}c@{}}\textbf{GPT-based}\\ \textbf{Eval.}\end{tabular}} & \multirow{2}{*}{\begin{tabular}[c]{@{}c@{}}\textbf{Human}\\ \textbf{Eval.}\end{tabular}} & \multicolumn{3}{c}{\textbf{Correlation}}      \\ \cmidrule{5-7} 
                             &                         &                                                                            &                                                                        & \#1 \& \#2 & \#1 \& \#3 & \#2 \& \#3 \\ \midrule
GPT-4V + mRAG       & 40.23                   & 37.50                                                                      & 42.00                                                                  & 0.47       & 0.34       & 0.66       \\
Qwen-VL-Chat + mRAG & 28.46                   & 26.50                                                                      & 25.00                                                                  & 0.49       & 0.41       & 0.69        \\
OmniSearch (G)          & 57.56                   & 54.00                                                                      & 49.00                                                                  & 0.45       & 0.38       & 0.63       \\
OmniSearch (Q)              & 41.39                   & 40.50                                                                      & 39.50                                                                  & 0.43       & 0.37       & 0.61     \\ \bottomrule  
\end{tabular}}
\label{tab:metric-corr}
\end{table*}

\noindent\textbf{Is more retrieved content useful?} Table \ref{tab:exp_topk} explores the impact of varying amounts of retrieved content on model performance, from which we can find that:

(1) The model incorporating retrieved content consistently surpasses its counterpart without retrieval, affirming the inherent advantage of mRAG.

(2) Unlike GPT-4V, which does not exhibit continual improvement with increased retrieval volume, our OmniSearch demonstrates superior capacity for utilizing extensive retrieved content. This indicates that despite potential noise contained in the retrieved content that may be harmful to question solving, OmniSearch can effectively filters or disregards such disturbances, thereby adequately leveraging the complex and voluminous retrieved content.

(3)  For the English question, the performance of both models continues to grow with the increase of retrieved content. This is partly due to the fact that both models use GPT-4V as the backbone, which is inherently more capable in English than in Chinese, and also arises from the fact that Google search is naturally more inclined towards English websites, leading to better support for English questions. This inspires us that in the future, more search tools can be introduce based on language characteristics, such as Bing, Baidu, etc., and even multiple search tools can be utilized to verify or vote for the final answers.

\subsection{Consistency of Different Evaluation Metrics}
In the main experiment, we employed F1-Recall as an evaluation metric due to its convenience. To demonstrate that it reliably reflects the true capabilities of models, we introduced two supplementary metrics: GPT-based Accuracy and Human-based Accuracy. For these metrics, we presented questions with ground-truth answers to GPT-4V and human evaluators, respectively, asking them to assess the correctness of the model responses and then compute the percentage of correct answers. Table \ref{tab:metric-corr} delineates the scores of the different models across these three metrics, as well as their correlation, which is quantified by the Pearson correlation coefficient. The Pearson coefficient ranges from -1 to 1, with 1 signifies a perfect positive correlation and -1 denotes a complete negative correlation. The trends of the different models across these three metrics are entirely consistent, with all coefficients exceeding 0, affirming a positive correlation. This demonstrates that F1-Recall fully reflects model performance. While GPT-based Accuracy and Human-based Accuracy exhibit stronger consistency, which may prove that they are more reliable, F1-Recall remains advantageous as an automated metric, offering significantly lower computational costs and better scalability.

\subsection{Supplementary Analysis of Computational Costs}

In Table \ref{tab:sup_token_cost}, we further reported the computational cost of OmniSearch on questions with different answer update frequencies (fast, slow, never). The results indicate that:

(1) Overall, OmniSearch consumes more tokens for more complex questions, such as those with fast or slow answer updates, primarily because these questions inherently require more retrieval steps.

(2) The difference in resource consumption for questions of varying difficulty is more pronounced for the smaller model OmniSearch (Q). This is due to the behavioral differences between GPT-4V and Qwen-VL. Specifically, as described in the Section \ref{sec:domain_perf}, GPT-4V tends to be more rigorous in question-solving, proactively planning verification retrievals to validate final answers, leading to retrieval processes exceeding three steps even for some relatively easier questions.

(3) By comparing rows 1 and 2 in the Table \ref{tab:sup_token_cost}, we can observe that after replacing OmniSearch (G)'s Sub-question Solver with a smaller Qwen-VL-Chat, the total token consumption of Qwen-VL-Chat and GPT-4V is stil comparable to the original model. This suggests that OmniSearch (G) effectively offloads computational burden to smaller models without a significant drop in overall performance.

\begin{table}[]
\centering
\renewcommand\arraystretch{1}
\setlength\tabcolsep{3.3pt}
\caption{The token cost of OmniSearch on questions with different answer update frequencies.}
\scalebox{0.65}{
\begin{tabular}{llccccccccc}
\toprule
\multirow{2}{*}{\textbf{Planning Model}} & \multirow{2}{*}{\begin{tabular}[c]{@{}l@{}}\textbf{Sub-question}\\ \textbf{Solver}\end{tabular}} & \multicolumn{3}{c}{\textbf{Fast}}                                                                                                                                     & \multicolumn{3}{c}{\textbf{Slow}}                                                                                                                                     & \multicolumn{3}{c}{\textbf{Never}}                                                                                                                                    \\ \cmidrule(lr){3-5} \cmidrule(lr){6-8} \cmidrule(lr){9-11}
                                &                                      & \textbf{Input T.}                                               & \textbf{Output T.}                                            & \textbf{Perf.} & \textbf{Input T.}                                               & \textbf{Output T.}                                            & \textbf{Perf.} & \textbf{Input T.}                                               & \textbf{Output T.}                                            & \textbf{Perf.} \\ \midrule
OmniSearch (G)                  & OmniSearch (G)                       & 3098.5 (G)                                                            & 466.3 (G)                                                            & 44.04               & 3036.5 (G)                                                             & 477.0 (G)                                                            & 49.58                & 2974.6 (G)                                                            & 483.9 (G)                                                           & 54.45                \\ \midrule
OmniSearch (G)                  & Qwen-VL-Chat                         & \begin{tabular}[c]{@{}c@{}}1268.2 (G)\\  + 2165.4 (Q)\end{tabular} & \begin{tabular}[c]{@{}c@{}}403.1 (G) \\ + 148.4 (Q)\end{tabular} & 38.65                & \begin{tabular}[c]{@{}c@{}}1214.1 (G) \\ + 1925.7 (Q)\end{tabular} & \begin{tabular}[c]{@{}c@{}}358.5 (G) \\ + 112.7 (Q)\end{tabular} & 44.68                & \begin{tabular}[c]{@{}c@{}}1185.6 (G) \\ + 2138.8 (Q)\end{tabular} & \begin{tabular}[c]{@{}c@{}}398.2 (G) \\ + 119.3 (Q)\end{tabular} & 52.25                \\ \midrule
OmniSearch (Q)                  & OmniSearch (Q)                       & 10258.2 (Q)                                                           & 638.1 (Q)                                                           & 35.16                & 9874.8 (Q)                                                            & 634.2 (Q)                                                           & 40.89                & 8866.9 (Q)                                                            & 475.3 (Q)                                                           & 45.52                \\ \midrule
OmniSearch (Q)                  & GPT-4V                               & \begin{tabular}[c]{@{}c@{}}2508.8 (G) \\ + 933.0 (Q)\end{tabular}  & \begin{tabular}[c]{@{}c@{}}269.3 (G) \\ + 547.8 (Q)\end{tabular} & 37.14                & \begin{tabular}[c]{@{}c@{}}2452.5 (G) \\ + 1000.9 (Q)\end{tabular} & \begin{tabular}[c]{@{}c@{}}233.6 (G) \\ + 490.5 (Q)\end{tabular} & 42.82                & \begin{tabular}[c]{@{}c@{}}2209.6 (G) \\ + 1024 (Q)\end{tabular}   & \begin{tabular}[c]{@{}c@{}}329.7 (G) \\ + 606.3 (Q)\end{tabular} & 47.48                \\ \bottomrule
\end{tabular}}
\label{tab:sup_token_cost}
\end{table}

Additionally, we also reported the average latency for each question in Table \ref{tab:sup_time_cost}. The results show that substituting some modules in OmniSearch with smaller models effectively reduces latency. Moreover, the ratio of search time to model inference time is roughly 2:3, indicating significant potential for optimization in both aspects. It is important to note that latency is a complex system engineering issue that involves not only the model's complexity but also factors network configuration of search APIs, caching strategies for retrieval content, inference model acceleration, and hardware FLOPS, etc.

\begin{table}[]
\centering
\caption{The average model latency for question in Dyn-VQA.}
\scalebox{0.8}{
\begin{tabular}{llccc}
\toprule
\textbf{Planning Model} & \textbf{Sub-question Solver} & \textbf{Search Time (s/\%)} & \textbf{Inference Time (s/\%)} & \textbf{Total Time (s)} \\ \midrule
OmniSearch (G)          & OmniSearch (G)               & 14.1 (44.8\%)               & 18.4 (55.2\%)                  & 31.5                    \\
OmniSearch (G)          & Qwen-VL-Chat                 & 12.3 (44.6\%)               & 15.3 (55.4\%)                  & 27.6                    \\
OmniSearch (Q)          & OmniSearch (Q)               & 8.5 (38.3\%)                & 13.7 (61.7\%)                  & 22.2                    \\
OmniSearch (Q)          & GPT-4V                       & 11.8 (45.0\%)               & 14.4 (55.0\%)                  & 26.2                    \\ \bottomrule
\end{tabular}}
\label{tab:sup_time_cost}
\end{table}

\begin{table}[t]
\centering
\caption{Model performance comparison on different VQA datasets. Heuristic mRAG in Table represents image caption-based mARG.}
\scalebox{0.8}{
\begin{tabular}{lcccc}
\toprule
\textbf{Model}                    & \textbf{VQAv2} & \textbf{A-OKVQA} & \textbf{InfoSeek} & \textbf{Dyn-VQA}\\ \midrule
\multicolumn{5}{l}{\textbf{\textit{Original MLLMs} }}                           \\
GPT-4V                   & 68.00       & 83.63        & 70.44 & 30.25        \\ \midrule
\multicolumn{5}{l}{\textbf{\textit{+ Heuristic Two-Step mRAG} }}                              \\
GPT-4V                   & 65.36      & 81.00        & 58.64   & 37.15      \\ \midrule
\multicolumn{5}{l}{\textbf{\textit{Ours} }}                              \\
OmniSearch (G)       & 70.34      & 84.12        &  71.48  & 50.03        \\ \bottomrule
\end{tabular}}
\label{tab:performance_other_datasets}
\end{table}

\section{Analysis Experiments on Dyn-VQA Dataset}
\subsection{Model Performance on Other VQA Datasets}
Table \ref{tab:performance_other_datasets} presents the model performances on various VQA datasets, highlighting several observations:

(1) The original GPT-4V achieves an average performance exceeding 74 on previous datasets, approaching human-level proficiency in the Table \ref{tab:difficult}. Conversely, its performance poorly on our Dyn-VQA with a significantly lower F1-reacll of 30.25, substantially lagging behind human capabilities. This discrepancy primarily arises because MLLMs have internalized much of the knowledge pertinent to traditional VQA datasets, where many questions rely on common-sense knowledge. For instance, VQAv2 frequently poses questions about object properties or intentions behind actions, which are relatively specialized by GPT-4V.

(2) The performance of the heuristic mRAG method is unstable on different types of questions (datasets). It impairs the effectiveness of GPT-4V on all previous datasets, especially on the InfoSeek where the decline is over 10 points. Through case analysis, we found that the heuristic mRAG struggles predominantly with questions of images depicting buildings, specific flora and fauna, which the image caption model fails to describe accurately. Consequently, the search engine yields ``shallow'' knowledge, i.e., content that is relevant to the question topic but is actually irrelevant to question, and instead misleads the original model. Moreover, given that InfoSeek is automatically generated from Wikipedia and characterized by homogenous question types, a substantial proportion of these problematic questions magnifies the disability of heuristic mRAG. On the contrary, the questions of VQAv2 and A-OKVQA typically inquire about common-sense knowledge, which is quite different from the real-world knowledge on the Internet, therefore the retrieved content instead has a minor (but still present) negative impact on the model.

(3) Our OmniSearch method achieved steady gains on each dataset. Even for datasets such as VQAv2 and A-OKVQA, which demand less extensive real-world knowledge from the Internet, OmniSearch still achieved slight growth. Since OmniSearch potentially makes search determinations, allowing it to avoid unnecessary retrieval interference with the intrinsic model understanding of questions that actually do not benefit from external knowledge augmentation. OmniSearch proves more adaptable and robust to diverse question types compared to the heuristic mRAG.

\begin{figure*}[th]
\centering
\scalebox{1}{
\includegraphics[width=0.98\textwidth]{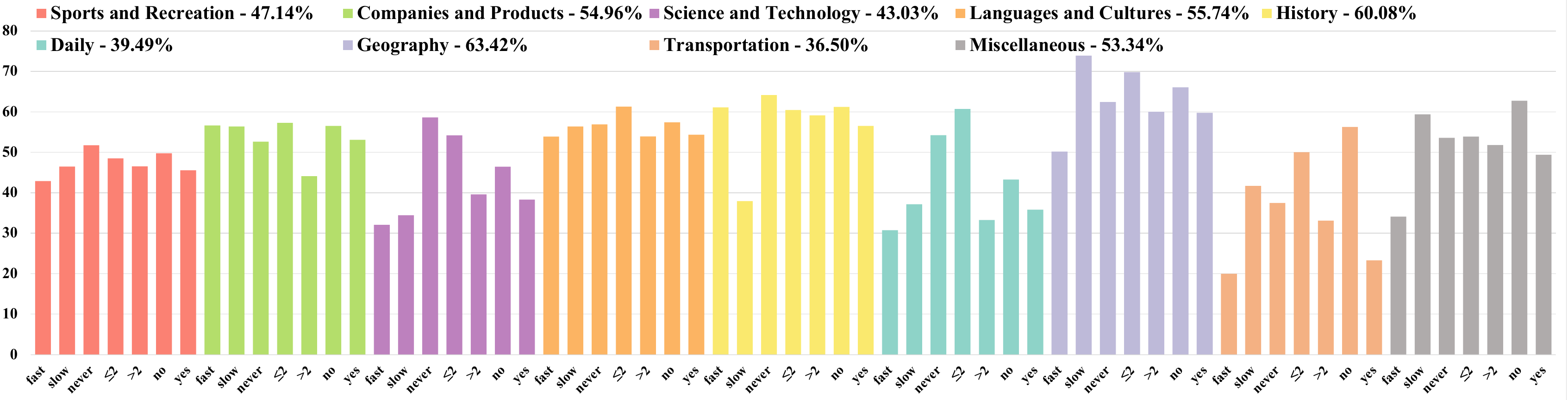}}
\caption{The performance of OmniSearch on different domains. Each category is represented by 7 bars of the same color, representing questions of different categories (in order fast, slow, never, $\leq$2-hop, $>$2-hop, no, yes).}
\label{fig:data_domain_2}
\end{figure*}
\subsection{OmniSearch Performance on Different Domains}
Figure \ref{fig:data_domain_2} displays the performance of OmniSearch across various domains, as an complement to Figure \ref{fig:radar} in the main content. We find that in most domains, the heights of the bars follow a similar pattern: 1st bar (fast) $<$ 2nd bar (slow) $<$ 3rd bar (never), 4th bar (less than 2-hop) $>$ 5th bar (more than 2-hop), 6th bar (not require external visual knowledge) $>$ 7th bar (require external visual knowledge). This indicates that OmniSearch generally underperforms on questions necessitating more complex retrieval processes. There are exceptions in data-sparse domains such as Transportation, where OmniSearch also shows relatively poor performs in these domains.

\begin{figure*}[t]
\centering
\scalebox{1}{
\includegraphics[width=0.98\textwidth]{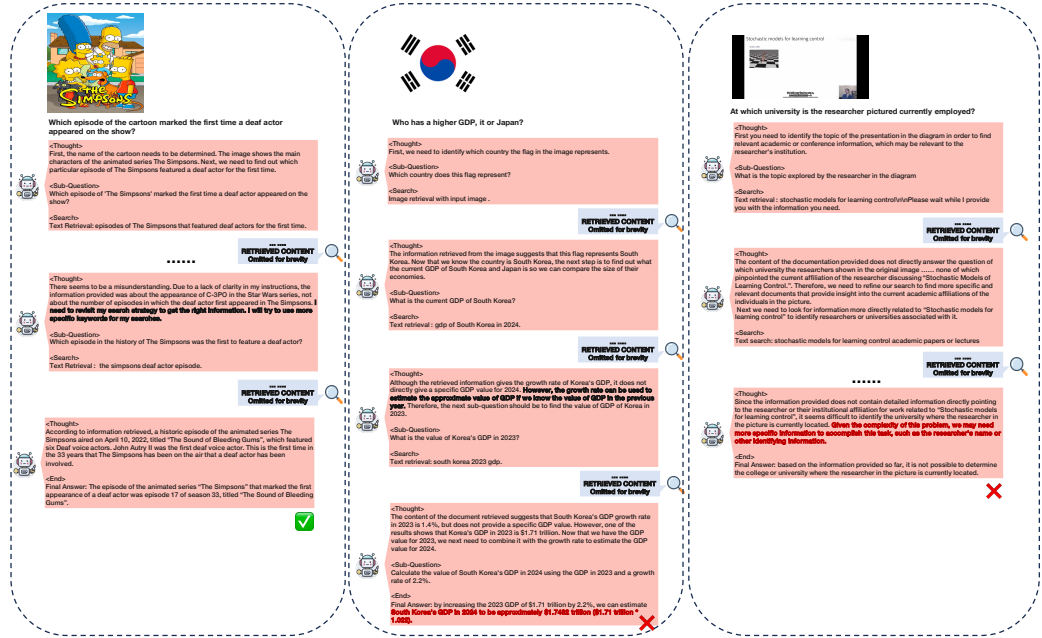}}
\caption{Case study with successful case and two bad cases.}
\label{fig:case}
\end{figure*}

\subsection{Case Study}
Figure \ref{fig:case} presents several intriguing case studies. In the first case on the left, the model dynamically adjusted its retrieval strategy multiple times after an initial retrieval failure, demonstrating OmniSearch's ability to adapt to complex retrieval environments and actively adjust its strategy based on environmental feedback. However, the subsequent two cases highlight different challenges faced by OmniSearch. In the middle example, after the model estimated the current GDP of South Korea by changing the direct search strategy to an indirect one with the growth rate and the previous year's GDP, it should have proceeded to retrieve Japan's current GDP. Instead, OmniSearch abruptly ceased its thought and retrieval. This indicates that OmniSearch struggles with questions requiring extended reasoning chains and tends to lose track of the original question and preceding information. In the right case, the initial planned retrieval strategy is incorrect from the start, as the main body of the image is a PowerPoint slice, and the person involved in the question only occupies a small space in the bottom right corner of the image. The OmniSearch focuses on the wrong visual evidence and gets caught in a "thinking trap". Ideally, the model should first search the image to find out that it is a screenshot of a course video, then view the video to find out the name of the speaker, and further search the web page to find out information about his academic institution. Another alternative is to get the exact search region, i.e., the bottom-right corner of the image, through image object recognition and image cropping. Then the person information is obtained based on the caption of the retrieved image. However, both approaches cannot be perfectly achieved by the current OmniSearch, which does not support such complex and fine-grained retrieval. These failed cases bring us significant insights and inspirations: firstly, how to solve the question requiring long context knowledge is worth studying, we analysed 100 error cases, and found that 73 of them encounter the issue of partially containing the correct answer, but the OmniSearch can't complete the full retrieval process due to the long context. On the one hand, we need to improve the maximum length of the context window of MLLMs, on the other hand, how to denoise, compress, and summarize the context is the direction that the sub-problem solver of OmniSearch can be improved. Secondly, advancing more precise retrieval techniques and incorporating a broader range of retrieval tools is an urgent research to be carried out.

\end{document}